\documentclass[11pt]{article}

\usepackage[utf8]{inputenc}
\usepackage[T1]{fontenc}
\usepackage{lmodern}
\usepackage{amsmath, amssymb, amsthm}
\usepackage{mathtools}
\usepackage{hyperref}
\usepackage{algorithm,algpseudocode}
\usepackage{graphicx}
\usepackage{float}
\usepackage{subcaption}
\usepackage{enumitem}
\usepackage{booktabs}
\usepackage{array}

\usepackage[margin=1in]{geometry}
\usepackage{setspace}
\usepackage{natbib}
\setcitestyle{numbers,sort&compress}

\hypersetup{
    colorlinks=true,
    linkcolor=blue,
    filecolor=magenta,      
    urlcolor=blue,
    citecolor=blue,
    pdfauthor={Alastair Poole},
    pdftitle={Projective Kolmogorov Arnold Neural Networks (P-KANs)},
    pdfsubject={Machine Learning, Neural Networks, Kolmogorov-Arnold Networks},
    pdfkeywords={Kolmogorov-Arnold Networks, Functional Spaces, Entropy minimisation, Signal Analysis, Dictionary Learning}
}

\title{\textbf{Projective Kolmogorov Arnold Neural Networks (P-KANs): \\Entropy-Driven Functional Space Discovery for Interpretable Machine Learning}}

\author{
    Alastair Poole$^{1}$, Stig McArthur$^{2}$, and Saravan Kumar$^{1}$\\[0.5em]
    $^{1}$ Centre for Applied Industrial Intelligence, University of Strathclyde \\
    $^{2}$ Design, Manufacturing, and Engineering Management, University of Strathclyde \\[0.5em]
    Contact: \texttt{alastair.poole@strath.ac.uk}
}

\date{\today}

\begin{document}

\maketitle

\begin{abstract}
Kolmogorov-Arnold Networks (KANs) are a revolutionary neural architecture design, relocating learnable nonlinearities from nodes to edges, demonstrating remarkable capabilities in scientific machine learning and interpretable modeling.
However, current KAN implementations suffer from fundamental inefficiencies due to redundancy in high-dimensional spline parameter spaces, where numerous distinct parameterizations yield functionally equivalent behaviors. This redundancy manifests as a "nuisance space" in the model's Jacobian, leading to a susceptibility to overfit and lose the potential to generalise beyond the training data.

We introduce Projective Kolmogorov-Arnold Networks (P-KANs), a novel training framework that guides edge function discovery towards interpretable functional representations through entropy-minimisation techniques from signal analysis and sparse dictionary learning.
Rather than constraining functions to predetermined spaces, our approach maintains spline space flexibility while introducing 'gravitational' terms that encourage convergence towards optimal functional representations, harnessing the universal approximation properties of splines.
Our key insight recognizes that optimal representations can be identified through entropy analysis of projection coefficients, compressing edge functions to lower-parameter projective spaces (Fourier, Chebyshev, Bessel in this instance).

P-KANs demonstrate superior performance across multiple domains, achieving up to 80\% parameter reduction while maintaining representational capacity, significantly improved robustness to noise compared to standard KANs, and successful application to industrial automated fiber placement prediction. Our approach enables automatic discovery of mixed functional representations where different edges converge to different optimal spaces based on local functional relationships, providing both compression benefits and enhanced interpretability for scientific machine learning applications.
While there are a limited number of projective spaces within this paper relying on pre-existing GPU accelerated data representations), it provides a doorway through which further abstracted structures can be placed on the functional space, validating the perspective that we can train both the parameters of the space, but also the functional space itself.

\end{abstract}


\section{Introduction}

Kolmogorov-Arnold Networks (KANs) have introduced a paradigm shift in neural architecture design by relocating learnable nonlinearities from nodes to edges, demonstrating remarkable capabilities in scientific machine learning and interpretable modelling~\cite{liu2025kankolmogorovarnoldnetworks}. Unlike traditional multilayer perceptrons with fixed nodal activation functions, KANs parameterise each edge with trainable functions, typically B-splines, that adapt to capture complex functional relationships with fewer parameters whilst maintaining interpretability. This innovation proves particularly valuable in domains requiring post-hoc rationalisation, such as Physics Informed Neural Networks (PINNs) and symbolic regression.

However, current KAN implementations introduce fundamental inefficiencies that limit compressed model deployment and robust generalisation. Standard architectures utilise covering spline spaces of degree $k$ with $p$ control points, creating high-dimensional parameter spaces where numerous distinct parameterisations yield functionally equivalent behaviours. This redundancy manifests as a 'nuisance space' in the model's Jacobian, where parameter perturbations produce negligible output effects. Whilst enabling universal approximation, this flexibility introduces overfitting susceptibility and demands unnecessarily large training datasets.

The core limitation stems from the disconnect between spline space structure and underlying functional relationships in real-world data. Many physical processes exhibit behaviours efficiently represented within established functional vector spaces: Fourier, Chebyshev, or wavelet domains, yet standard KAN training provides no mechanism to discover these natural representations. Consequently, edge functions converge to unnecessarily complex spline configurations that could be more parsimoniously expressed using appropriate basis functions.

We address this limitation by introducing Projective Kolmogorov-Arnold Networks (P-KANs), a training framework that guides edge discovery towards interpretable functional representations through entropy-minimisation techniques from signal analysis and sparse dictionary learning. Rather than constraining functions to predetermined spaces such as Wav-KANs \cite{bozorgasl2024wavkanwaveletkolmogorovarnoldnetworks} or Fourier-KANs \cite{zhang2025kolmogorovarnoldfouriernetworks}, our approach maintains spline space flexibility whilst introducing 'gravitational' terms encouraging convergence towards optimal functional representations.

Our key insight recognises that optimal representations can be identified through entropy analysis of projection coefficients when edge functions decompose into candidate spaces. By computing discrete transforms into multiple inner-product spaces (Fourier, Chebyshev, Bessel), we quantify representation efficiency and dynamically guide optimisation towards suitable functional forms. This enables automatic discovery of mixed functional representations where different edges converge to different optimal spaces.

\subsection{Contributions to Knowledge}

\begin{enumerate}[label = (\roman*)]
    \item Unified functional framework: Mathematical treatment of multiple orthonormal inner-product spaces within a single optimisation scheme using entropy-based metrics within the field of KANs.
    \item Entropy-driven training: First systematic approach for guiding KAN edge functions towards structured representations through differentiable entropy minimisation without \textit{a priori} knowledge of underlying relationships.
    \item Adaptive regret algorithm: Comprehensive training algorithm with regret functions that dynamically revert inappropriate assignments, ensuring robust convergence when initial selections prove suboptimal.
\end{enumerate}

The remainder proceeds as follows: Section 2 surveys related work in functional representations, dictionary learning, and KAN advances; Section 3 presents our methodology and complete algorithm; Section 4 provides experimental validation demonstrating efficiency gains and interpretability benefits; Section 5 discusses implications and future research directions.

\section{Related Work}

\subsection{Kolmogorov-Arnold Neural Networks}

Kolmogorov-Arnold Neural Networks emerged from the foundational work of Liu et al.~\cite{liu2025kankolmogorovarnoldnetworks}, drawing inspiration from the celebrated Kolmogorov-Arnold representation theorem~\cite{kolmogorov1957representation, Arnold2009}. This architecture fundamentally departs from classical multilayer perceptrons by relocating learnable nonlinearities from nodes to edges, where each connection implements a trainable function—typically a spline—that transforms signals between neurons.

The mathematical foundation of KANs rests upon the universal approximation properties of splines. For any function $f \in C^{k+1}[a, b]$ that is continuous up to its $k^{\text{th}}$ derivative, there exists a spline $s$ of degree $k$ on a sufficiently fine knot partition such that
\begin{equation}
    |f - s|_{\infty} \leq C h^{k+1}\left|f^{(k+1)}\right|_{\infty},
\end{equation}
where $h$ represents the maximum knot spacing and $C$ depends only on $k$~\cite{deboor2001practical, Schumaker_2007}. This theoretical guarantee ensures that spline-parameterised KANs achieve density in the space of continuous functions, enabling them to model the broad range of relationships encountered in real-world applications.

A single KAN layer can be expressed mathematically as:
\begin{equation}
    y_j = \sum_{i=1}^{n} f_{ij}(x_i) + b_j
\end{equation}
where $f_{ij}$ denotes the learned edge spline connecting input $x_i$ to output node $j$, and $b_j$ represents the bias term. This edge-centric architecture distinguishes KANs from traditional neural networks, as the network's expressivity emerges from the space of edge functions rather than from node-wise nonlinearities.

KANs have demonstrated particular efficacy in scientific machine learning applications, where their interpretability proves crucial. The spline-based edge functions enable post-hoc extraction of symbolic relationships, whilst the reduction in parameter count (in comparison to traditional neural networks of equivalent capacity) facilitates training on smaller datasets. However, these advantages come at the cost of potential redundancy within the spline parameter space, a limitation that motivates the present work.

\subsection{Signal Analysis and Dictionary Learning}

The field of signal analysis has long recognised the value of representing signals as sparse linear combinations of basis functions drawn from structured dictionaries. This approach enables both compression and interpretability by decomposing complex signals into fundamental components that correspond to well-understood mathematical objects.

\subsubsection{Sparse Dictionary Creation}

Classical sparse dictionary learning seeks to represent a signal $\mathcal{D}$ over domain $x$ as a sparse linear combination of basis functions. The fundamental optimisation problem takes the form:
\begin{equation}
    \min_{\beta \in \mathbb{R}^{n}} \left\| \mathcal{D} - A\beta\right\|^{2}
\end{equation}
where $n$ denotes the dictionary size and $A$ represents the matrix encoding of dictionary elements over $x$.

Regularisation techniques extend this framework to balance reconstruction accuracy against desirable properties such as sparsity or smoothness. The Lasso method introduces an $L_1$ penalty:
\begin{equation}
    \min_{\beta \in \mathbb{R}^{n}} \left\| \mathcal{D} - A\beta\right\|^{2} + \lambda \|\beta\|_1
\end{equation}
whilst Ridge regression employs an $L_2$ penalty:
\begin{equation}
    \min_{\beta \in \mathbb{R}^{n}} \left\| \mathcal{D} - A\beta\right\|^{2} + \lambda \|\beta\|_2^2
\end{equation}

The Lasso formulation encourages sparsity through its $L_1$ norm, proving particularly valuable for feature selection and preventing overfitting in noisy conditions. Ridge regression, conversely, promotes mixed dictionary usage whilst discouraging any single basis function from dominating the solution. ElasticNet combines both approaches, efficiently balancing sparsity with solution diversity to enhance retraining robustness.

These techniques find widespread application in non-destructive testing (NDT), statistical analysis, and numerous domains where data exhibit inherent sparsity or noise, and where the space of representative functions remains poorly understood. The artificial intelligence community has not overlooked these regularisation benefits, with Wen et al.~\cite{wen2016learningstructuredsparsitydeep} successfully applying structured sparsity to traditional neural network architectures through active learning dependencies on edge parameters.

\subsubsection{Functional Vector Spaces with Inner Products}

Many signal analysis applications employ orthonormal basis vectors that span specific functional vector spaces, defined by their mutual inner products equalling zero. Practical implementations often focus on single-space representations, either because the domain offers intuitive advantages or because empirical evidence demonstrates superior performance for specific problem classes.

Medical ultrasound exemplifies this approach, where Yousufi et al.~\cite{aplhagasjia} provide comprehensive analysis of sparse representation techniques, highlighting the trade-offs between various regimes with particular emphasis on wavelet and Fourier transforms. Computer vision research has similarly demonstrated that Chebyshev polynomial-based filters—another orthonormal vector space—deliver excellent reconstruction quality for noisy images~\cite{Tian_2014}.

Single-vector-space dictionaries prove most valuable when practitioners possess strong understanding of the problem domain or when the data follow specific patterns determined by particular differential equations. This domain knowledge enables cultivation of specialised functional approaches that exploit the underlying mathematical structure of the phenomena being modelled.

\subsection{Functional Representations in KAN Architectures}

Recent research has increasingly focused on applying specific functional bases to KAN edge representations, drawing parallels with the dictionary learning approaches established in signal analysis.

Bozorgasl and Chen introduced Wav-KANs~\cite{bozorgasl2024wavkanwaveletkolmogorovarnoldnetworks}, utilising mixed orthonormal and semi-orthonormal wavelet functions as edge function bases. Their results indicate superior performance, convergence, and accuracy compared to standard spline-based KANs, potentially due to the restricted number of active parameters during training whilst maintaining interpretable outputs through explicit functional representations. Similarly, the work of SS et al.~\cite{ss2024chebyshevpolynomialbasedkolmogorovarnoldnetworks} introduces Chebyshev-based edge activation functions, demonstrating potential for fast parallelised computations when optimising over intrinsically diagonalised orthonormal basis spaces.

Zhang et al.~\cite{zhang2025kolmogorovarnoldfouriernetworks} achieved remarkable results by replacing spline edge functions with Fourier series, significantly reducing the trainable parameter space whilst maintaining impressive performance across diverse problem domains. These approaches share common characteristics with the specialised sparse dictionary decomposition methods employed in signal analysis applications.

The trend across these investigations mirrors that observed in sparse dictionary decomposition: orthogonal or semi-orthogonal basis functions enable more efficient edge function training, with basis orthogonality facilitating faster convergence and improved generalisation. These approaches additionally provide interpretable outputs, as the edge functions operate in terms of well-established mathematical functions, simplifying model behaviour interpretation—particularly valuable in applications such as NDT, where edge functions require interpretation rather than potentially over-parameterised or complex data representations.

Alternative approaches have explored different functional forms. Li~\cite{li2024fastkan} and Ta~\cite{ta2024bsrbfkancombinationbsplinesradial} successfully applied Radial Basis Functions (RBFs) either independently or supplementing B-spline approaches. These developments diverge from other advances in the field, as RBFs do not constitute naturally orthogonal bases in the strictest sense, yet demonstrate potential for less stringent functional approaches.

Howard et al.~\cite{howard2024finitebasiskolmogorovarnoldnetworks} pursued an alternative direction, seeking separability in terms of domain rather than functional space. They generate topological atlases~\cite{jost2013riemannian} of representative KANs over input domains, reducing overall model complexity through intelligent training space selection. Whilst this approach differs from the functional space decomposition examined in the present work, it shares the spirit of mixed-domain decomposition explored herein.

Recent work has begun addressing scenarios where the correct basis representation remains unknown, necessitating naive approaches for efficient data representation learning and mixed-basis consideration. Aghaei~\cite{aghaei2024fkanfractionalkolmogorovarnoldnetworks} investigates generalising B-spline basis functions by ascending abstraction hierarchies, with promising applications to KANs. Ta et al.~\cite{ta2025fckanfunctioncombinationskolmogorovarnold} advance this direction by considering combinations of well-established functional vector spaces as B-spline activations.

However, existing approaches typically commit to specific functional spaces a priori or require manual specification of appropriate representations. None provide a systematic framework for automatically discovering optimal functional representations across multiple candidate spaces whilst maintaining the flexibility to revert inappropriate assignments during training. This limitation motivates our development of P-KANs, which extend these ideas by implementing a generalisable dictionary learning technique that automatically identifies efficient representations for datasets with poorly understood dynamics or mixed functional behaviours.

\section{Methodology}

\subsection{Problem Formulation}

Rather than constraining edge functions to predetermined functional spaces, we seek to construct a framework that automatically discovers efficient representations through dynamic space selection. This approach addresses a fundamental limitation in current KAN implementations: the redundancy inherent in high-dimensional spline parameter spaces, which manifests as numerous functionally equivalent parameterisations.

The redundancy of variables in a trained network can be quantified through singular value analysis of the Jacobian matrix. For a network $f(\theta, x)$ with hyperparameters $\theta_i$ and vector output, the Jacobian entry
\begin{equation}
    J_{i,j} = \frac{\partial f_i}{\partial \theta_j}
\end{equation}
quantifies the impact of hyperparameter $\theta_j$ on the $i^{\text{th}}$ output component after training.

Large numbers of small Jacobian entries indicate an overdetermined solution, intuitively representing fitting to noise or low-amplitude patterns. Oymak et al.~\cite{oymak2019generalization} characterise this phenomenon as a `nuisance space', demonstrating that avoiding training within these dimensions ensures model generalisability whilst achieving substantial training gains within the 'informative' hyperparameter subspace. This proves particularly valuable when working with noisy or incomplete datasets.

However, when discovering structural similarities between datasets through fine-tuning representative networks, greater local uniqueness assists in differentiating subtle relationships between models. This places specific requirements on the data: to understand a process comprehensively, one requires complete datasets. Industrial scenarios often satisfy this requirement through abundant sensing capabilities employed in systems monitoring and root-cause analysis. Crucially, we focus on tight data representation rather than out-of-distribution generalisation typical in machine learning applications~\cite{idnani2023dont}.

\begin{figure}[H]
  \centering
  \includegraphics[width=0.75\linewidth]{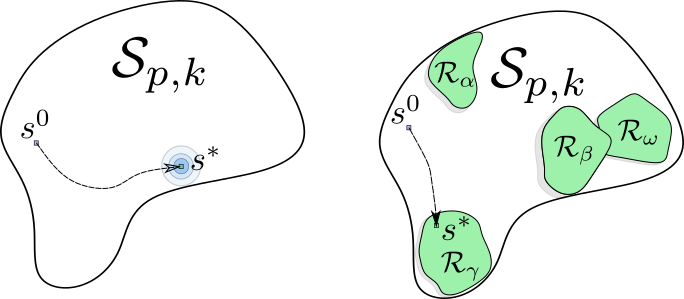}
  \caption{An edge spline is optimised within the functional space from its initial position $s^{0}$ to its final position $s^{*}$. Traditional KANs (left) optimise egde functions within the overarching spline-space determined by their degree ($k$) and number of spline points ($p$). When they have converged, the large number of parameters will lead to redundancies, represented in blue, where perturbations of the spline within the space will have little effect on the output function. The proposed method (right) recognises the approximal representation space of well known vector spaces (highlighted as shadows), and introduces a 'gravitational' term optimising towards these well known, lower-dimensional spaces.}
  \label{space_selection}
\end{figure}

We influence the distribution of nuisance values in the Jacobian by encouraging data representations towards locally unique, sparse, and flexibly discovered forms along each edge spline. Additionally, we can adjust the representative network size to promote healthy distributions of the Jacobian's singular value spectrum, either to identify highly influential features or to generate stable yet generalisable models - graphically visualised in \ref{space_selection}.

\subsection{Entropy-Driven Functional Space Projection}

Our approach draws upon the established principle that optimal basis selection minimises representation entropy. Coifman and Wickerhauser~\cite{119732} demonstrated that entropy-based algorithms provide systematic methods for identifying best basis representations in signal decomposition, establishing the theoretical foundation for our functional space selection mechanism.

We focus on vector spaces equipped with inner products that admit smooth transitions between basis elements, enabling both optimisation towards and within candidate spaces for each edge individually. This framework accommodates both finite and uncountably infinite bases whilst maintaining differentiability for gradient-based optimisation.

\subsubsection{Entropic Attraction to Inner Product Spaces}

We introduce a novel cost function based on the entropy of edge representations projected into candidate functional domains. Consider an edge spline $s_{l,i,j}(x)$ at layer $l$ connecting neurons $i \rightarrow j$. We project this function into an orthogonal functional vector space $\mathcal{R}$ (such as Fourier space):
\begin{equation}
    s(x) \rightarrow s^{r}(x) = \sum_q \alpha_q r^q(x)
\end{equation}
where $r^q$ represent the basis functions within the vector space. For notational convenience, we suppress layer and neuron indices.

We employ discrete formulations rather than continuous integrals to avoid computational burden, evaluating splines at discrete grid points with interpolation. All functional vector spaces operate discretely, with orthogonal spaces defined through discrete parameter sets obtained via transforms such as Discrete Fourier Transforms (DFT) or Discrete Wavelet Transforms (DWT).

Having effectively binned the data into a functional vector space with bin sizes determined by coefficient vector $\alpha_q$, we calculate the representation entropy by normalising the coefficient vector:
\begin{equation}
    \hat{\alpha}_q = \frac{|\alpha_q|}{\sum_t |\alpha_t|}
\end{equation}
and computing the entropy as:
\begin{equation}
    E_{\mathcal{R}} = -\sum_q \hat{\alpha}_q \log(\hat{\alpha}_q)
\end{equation}

This metric evaluates the efficiency of fitting the distribution within the specified functional vector space, following the approach established by Coifman and Wickerhauser~\cite{119732} for optimal basis identification in signal decomposition.

When considering $m$ spaces simultaneously, we obtain entropy values for different spaces $\mathcal{R}$:
\begin{equation}
    E_{\Sigma} = [E_{\mathcal{R}_0}, E_{\mathcal{R}_1}, \ldots, E_{\mathcal{R}_m}]
\end{equation}

To identify the optimal space for optimisation, we apply the softmin function to select the minimum entropy element:
\begin{equation}
    E^* = \frac{1}{\sum_k e^{-\lambda E_{\mathcal{R}_k}}} \sum_{g=0}^m E_{\mathcal{R}_g} e^{-\lambda E_{\mathcal{R}_g}}
\end{equation}
where the exponent terms are weighted by $\lambda$ to balance individual and collective minimisation. We begin with a small $\lambda$, progressively increasing the weight so that the random initialisation of the network is not biased towards specific spaces.

Each step in this process maintains smoothness, enabling differentiable backpropagation of the entropy metric. We could alternatively consider entropy of energy distributions using squared absolute coefficients, as employed by Zhuang and Baras~\cite{zhuang1994optimal} for optimal basis identification in multi-resolution wavelet decompositions. However, we restrict ourselves to absolute amplitude distributions to encourage sparsity within the discovered vector space rather than diversity of basis functions—analogous to Lasso and Ridge regression outcomes.

This approach effectively implements dynamic dictionary learning for activation functions, enabling the network to discover whether specific functional forms provide parsimonious relationship descriptions. For instance, if an edge function exhibits near-sinusoidal behaviour, the Fourier basis yields one large coefficient at the corresponding frequency with remaining coefficients near zero, producing very low entropy and encouraging training progression in that direction.

\subsection{Unified Cost Function}

The complete cost function combines standard loss minimisation with entropy-based functional space attraction and edge regularisation:
\begin{equation}\label{eq:full_cost}
    \begin{aligned}
        Q = \quad & \underbrace{\alpha \cdot \text{loss}}_{\text{reconstruction term}} 
         + \underbrace{\beta \cdot \overline{E^*_{l,i,j}}}_{\text{entropy minimisation}} 
         + \underbrace{\gamma \cdot \overline{\sum_{x \in \Omega_{l,i,j}} \frac{\|s_{l,i,j}(x)\|_{\ell^m}}{|\Omega_{l,i,j}|}}}_{\text{edge regularisation}}
    \end{aligned}
\end{equation}
where $\overline{\cdot}$ denotes averaging across all edges specified by layer and neuron indices, $\Omega_{l,i,j}$ represents sampling points along each edge, and $\ell^m$ indicates the chosen $m$-norm.

The reconstruction term ensures adherence to training objectives, the entropy term guides functional space discovery, and the regularisation term can promote either diversity or sparsity depending on the choice of norm and coefficient magnitude.

\subsection{Algorithm}

Our training process discovers efficient data representations whilst encouraging edge functions to converge towards specific functions within optimal functional spaces. Following initial training with the unified cost function, we identify best-fit spaces and functions by projecting splines into candidate spaces, determining major components, and applying the $R^2$ methodology employed by Liu et al.~\cite{liu2025kankolmogorovarnoldnetworks} with threshold relaxation across iterations.

After cost function optimisation, limited grid points may restrict spline shapes, potentially yielding poorly fitting functions. We address this through retraining cycles where fixed edges replace trainable spline knot parameters with simple parametric curves, fine-tuning the model to new functional topologies.

Given potential misidentification of correct functional bases, we implement regret functions tracking retraining loss to identify when to revert edges to spline representations, enabling retraining towards alternative basis spaces. Algorithm~\ref{alg:pkan} presents the complete workflow.

\begin{algorithm}[H]
    \caption{Projective KAN Training with Entropy-Driven Space Selection}
    \label{alg:pkan}
    \begin{algorithmic}
        \Require Training data $\mathcal{D}$
        \Require Initial KAN architecture with spline-based edges
        \Require Candidate function spaces $\mathcal{F} = \{\text{Fourier}, \text{Chebyshev}, \text{Bessel}, \ldots\}$
        \Require Loss coefficients $\alpha, \beta, \gamma$
        \Require Threshold $R^2_{\min}$ for basis substitution
        \Require Maximum training rounds $T$
        \Require Regret tolerance $\tau_{\text{regret}}$
        
        \State Initialise KAN model with spline-based edge functions
        \State Train model using unified cost function (Eq.~\ref{eq:full_cost})
        
        \For{$t = 1$ to $T$}
            \ForAll{edges $e$ in model}
                \State Project $e$'s spline into candidate functional spaces $\mathcal{F}$
                \State Compute $R^2$ score for best-fit function in each space
                \If{$\max R^2 > R^2_{\min}$}
                    \State Replace $e$'s spline with best-fit parametric function
                    \State Freeze edge and remove spline knot parameters
                \EndIf
            \EndFor
            
            \State Fine-tune model with fixed edge functions
            
            \ForAll{fixed edges $e$}
                \State Compute retraining loss and update regret function
                \If{regret exceeds $\tau_{\text{regret}}$}
                    \State Revert $e$ to spline-based representation
                    \State Unfreeze edge for continued training
                \EndIf
            \EndFor
            
            \State Reset optimiser to accommodate parameter changes
        \EndFor
        
        \State \Return trained model and discovered symbolic edge functions
    \end{algorithmic}
\end{algorithm}

\subsection{Implementation}

We implement P-KANs as PyTorch extensions building upon the pykan framework~\cite{pykan2025}, leveraging established foundations whilst introducing novel modules through the PyTorch backend. Our implementation supports both inner-product spaces (IPS) and general vector spaces (VS):

\textbf{Inner-Product Spaces:}
\begin{itemize}
    \item \textbf{Fourier space}: Implemented using PyTorch's built-in Discrete Fourier Transform.
    \item \textbf{Chebyshev polynomials}: Solutions to Chebyshev differential equations using the torch-dct package~\cite{hu2018torchdct} for coefficient identification.
    \item \textbf{Bessel functions}: Solutions to radial harmonic equations in positive domains using Gamma-function expansions with trigonometric approximations.
\end{itemize}


Each functional family was selected for belonging to well-established harmonic equations or representing distinct limiting behaviours in real analysis, ensuring broad coverage of functional relationships commonly encountered in scientific and engineering applications.

\section{Results and Analysis}\label{results}

Experimental validation of the P-KAN framework comprises three components: ablation studies examining individual contributions, robustness testing under noise conditions, and application to automated fibre placement manufacturing. Results quantify both theoretical advantages and practical limitations of entropy-driven functional projection.

\subsection{Ablation Study}\label{ablation}

The P-KAN architecture was tested under various conditions to validate component contributions. Test functions selected:
\begin{itemize}
  \item Simple sinusoidal (single sine function)
  \item Gaussian (exponential of square polynomial)
  \item Discontinuous (not smooth)
  \item Oscillatory (combination of sin and cos functions)
  \item Polynomial (cubic)
\end{itemize}

These represent a range of smoothnesses from none to infinite and varying complexity in each space's representation. No perfect B-spline or function space representation exists for discontinuous functions.

We want the data to be representative of the types we would see in real-world applications. Instead of generating data from a regular coordinate grid pattern, we sample data from random coordinates on the compact input subspaces.

\subsubsection{Model Complexity}\label{complexity_analysis}

Representational capability was measured using the $R^2$ score employed in the KAN toolbox:
\begin{equation}
  R^{2} = 1 - \frac{\sum_{i}\left( y_{i} - F(x_{i})\right)^2}{\sum_{i}\left( y_{i} - \bar{y}\right)^2},
\end{equation}
where $F(x)$ is the model's prediction.

\begin{figure}[H]
  \centering
  \includegraphics[width=0.75\linewidth]{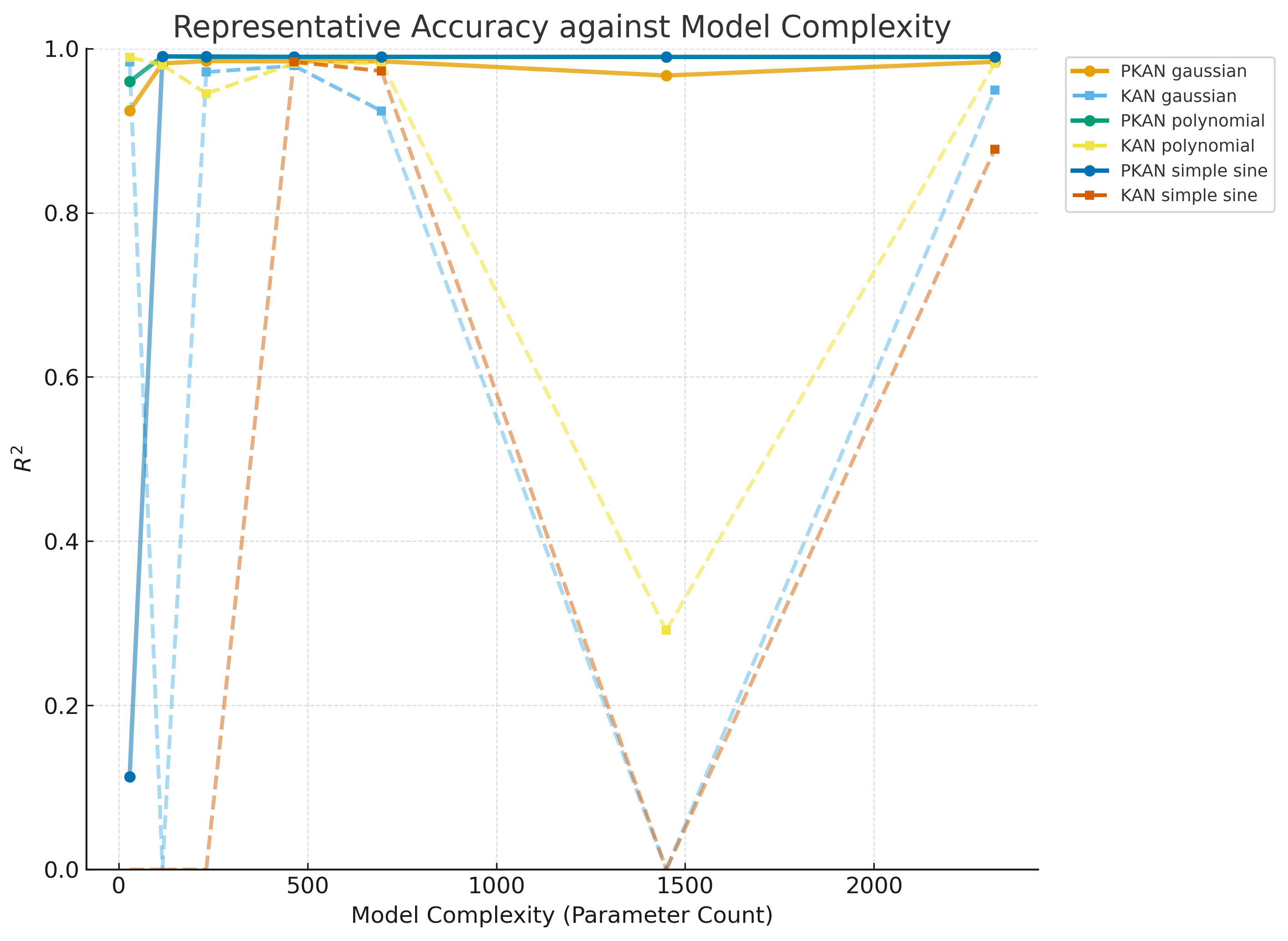}
  \caption{Base-complexity of each model against each model's representative capacity. While P-KAN produced a stable score, the KAN models do not converge as well.}
  \label{complexity_rsqd}
\end{figure}

Results in Figure \ref{complexity_rsqd} show P-KAN achieves high $R^2$ scores consistently across all but minimal model complexity. KAN models show instability in their results with larger models. For P-KAN models, baseline complexity is reduced: spline edges were cubic with 20 control points (n = 23 degrees of freedom per edge), whereas the optimised edge representation is reduced to n = 4 parameters, a reduction of over 80\%.

\begin{figure}[H]
  \centering
  \begin{minipage}{0.48\linewidth}
    \centering
    \includegraphics[width=\linewidth]{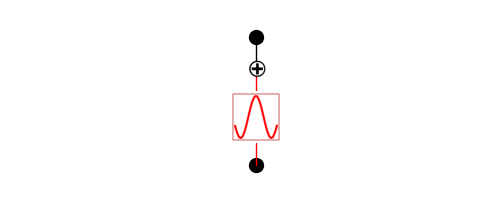}
    \subcaption{P-KAN Minimal}
  \end{minipage}
  \begin{minipage}{0.48\linewidth}
    \centering
    \includegraphics[width=\linewidth]{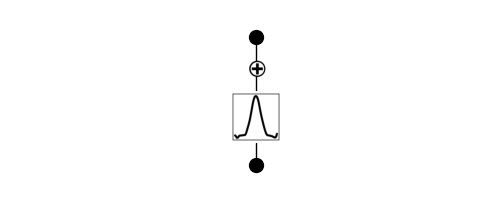}
    \subcaption{KAN Minimal}
  \end{minipage}
  \begin{minipage}{0.48\linewidth}
    \centering
    \includegraphics[width=\linewidth]{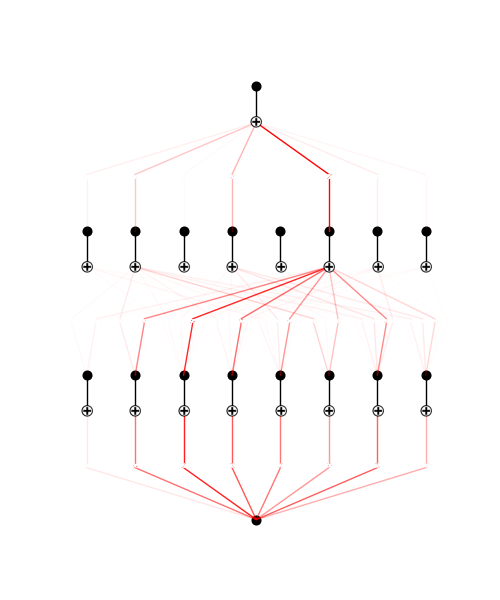}
    \subcaption{P-KAN Large}
  \end{minipage}
  \begin{minipage}{0.48\linewidth}
    \centering
    \includegraphics[width=\linewidth]{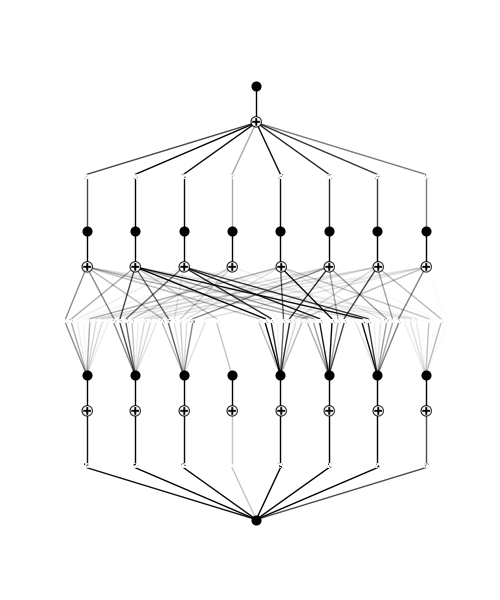}
    \subcaption{KAN Large}
  \end{minipage}  
  \caption{For Gaussian data, the architecture of P-KAN and KAN over different model complexities.}
  \label{arch_layout}
\end{figure}

This can be understood through P-KAN's parameter minimisation as the search space is significantly reduced, while demonstrating the same representative capabilities as the KAN network. A side-by-side comparison is shown in Figure \ref{arch_layout}.

\subsubsection{Functional Spaces}\label{functional_spaces}

\begin{figure}[H]
  \centering
  \includegraphics[width=0.75\linewidth]{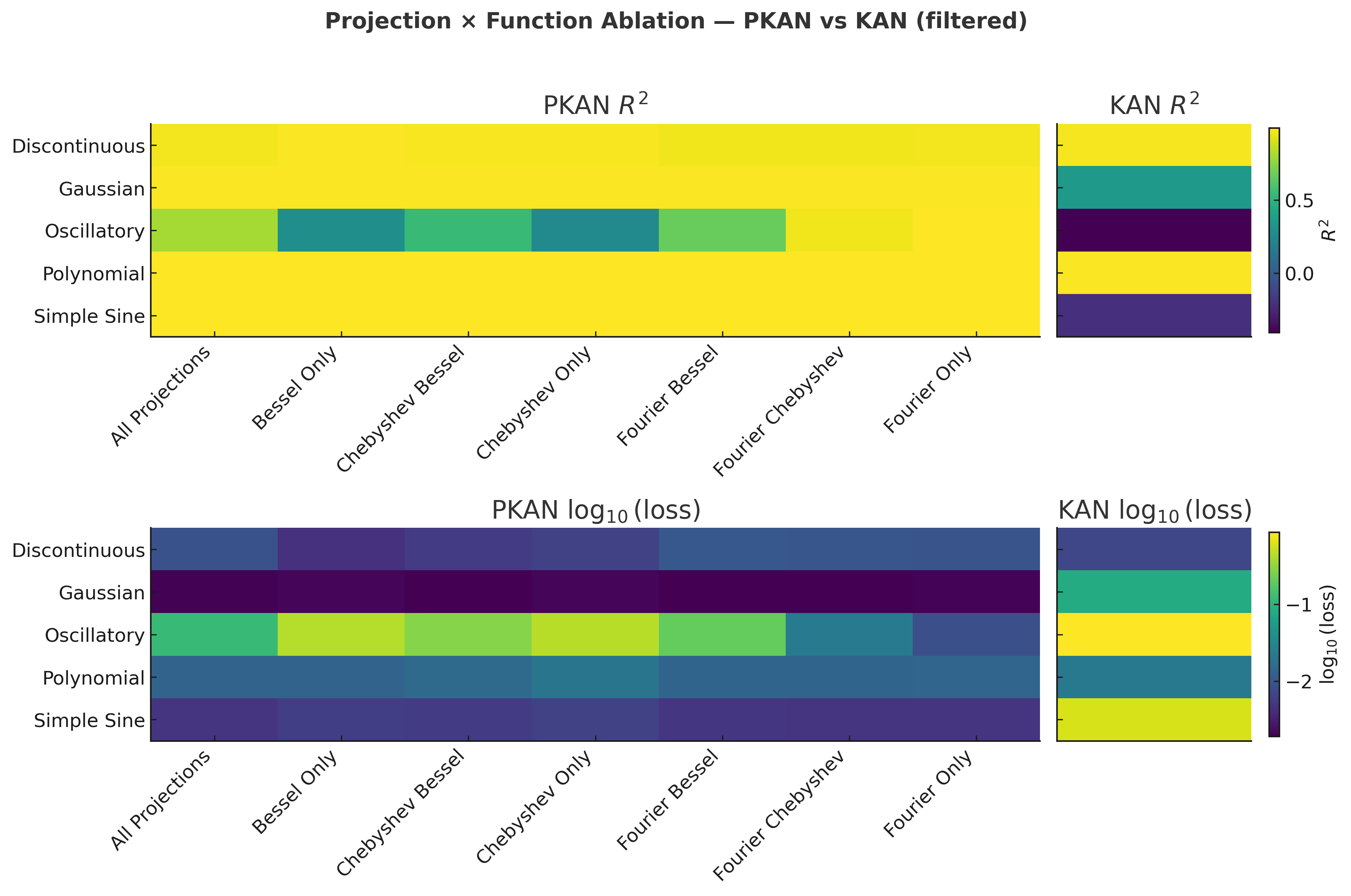}
  \caption{For each test function and projection combination, the training data $R^2$ score and log-loss is shown, with KAN acting as a reference on the right hand side. Each functional space is shown to be effective in some domain, with no single space being best across all domains.}
  \label{func_abl}
\end{figure}

Comparing efficacy of each functional space combination, Figure \ref{func_abl} shows each potential functional space is as good at or better than KAN results (that have outliers, Sine Polynomial and Discontinuous having 14\% outlier rates).

\begin{figure}[H]
  \centering
  \includegraphics[width=0.75\linewidth]{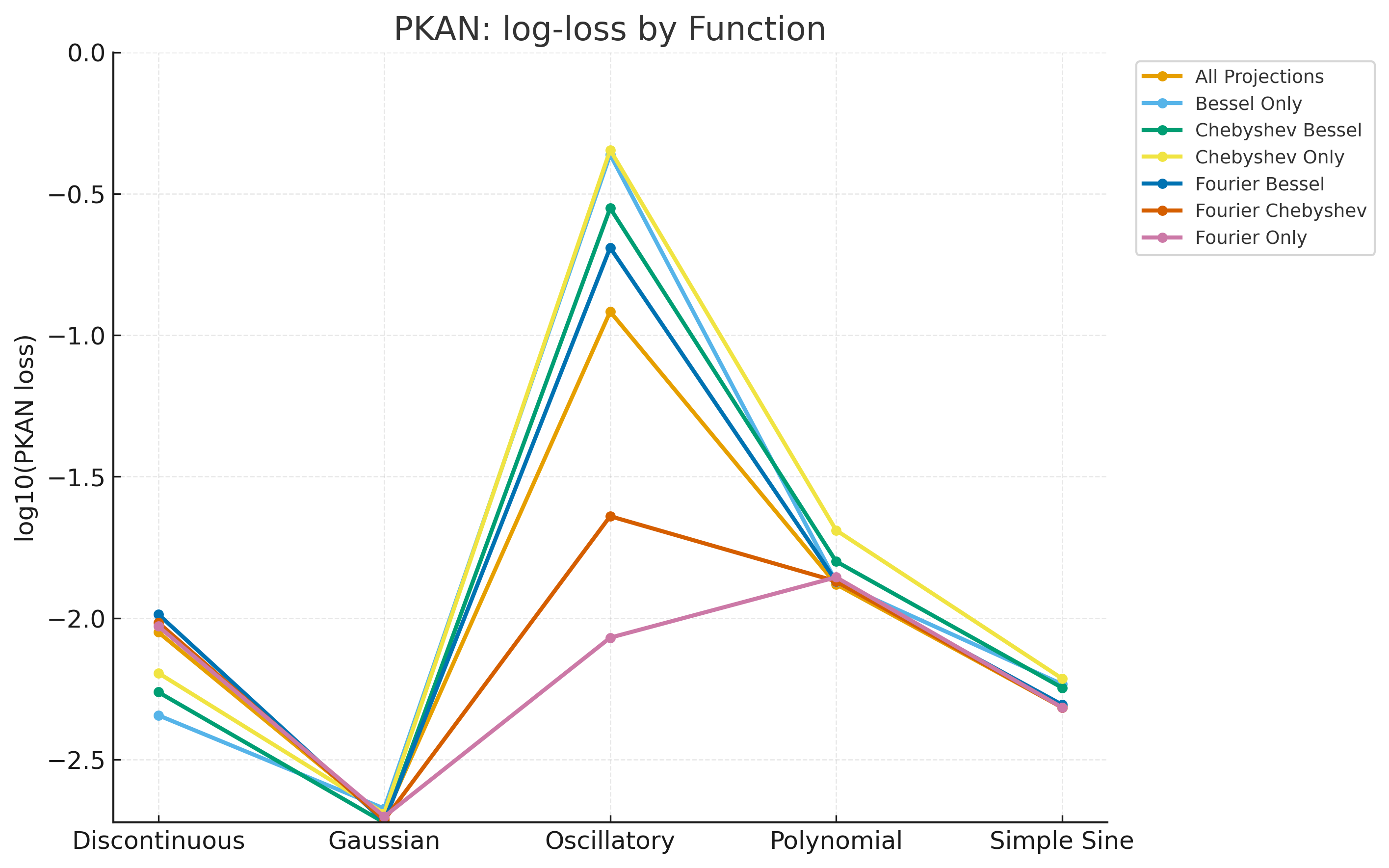}
  \caption{Comparison of the log-loss curves for each combination of functional spaces on the different test data sets. Each functional approach is shown to specialise in some problem, with P-KAN's naive approach providing a good all-rounder.}
  \label{fun_logloss}
\end{figure}

The only function P-KANs struggle with relatively is sine curves. However, viewing the log-loss curves in Figure \ref{fun_logloss}, we see that while all combinations converge relatively well, the P-KAN approach finds a suitable medium of all results. The combined framework can identify a good configuration without testing each architecture individually.

Further experimentation with the optimisation strategy is important, as stochastic approaches to functional space optimisation may provide better results.

\subsubsection{Spectral Analysis}\label{spectral_analysis}

\begin{figure}[H]
  \centering
  \includegraphics[width=0.75\linewidth]{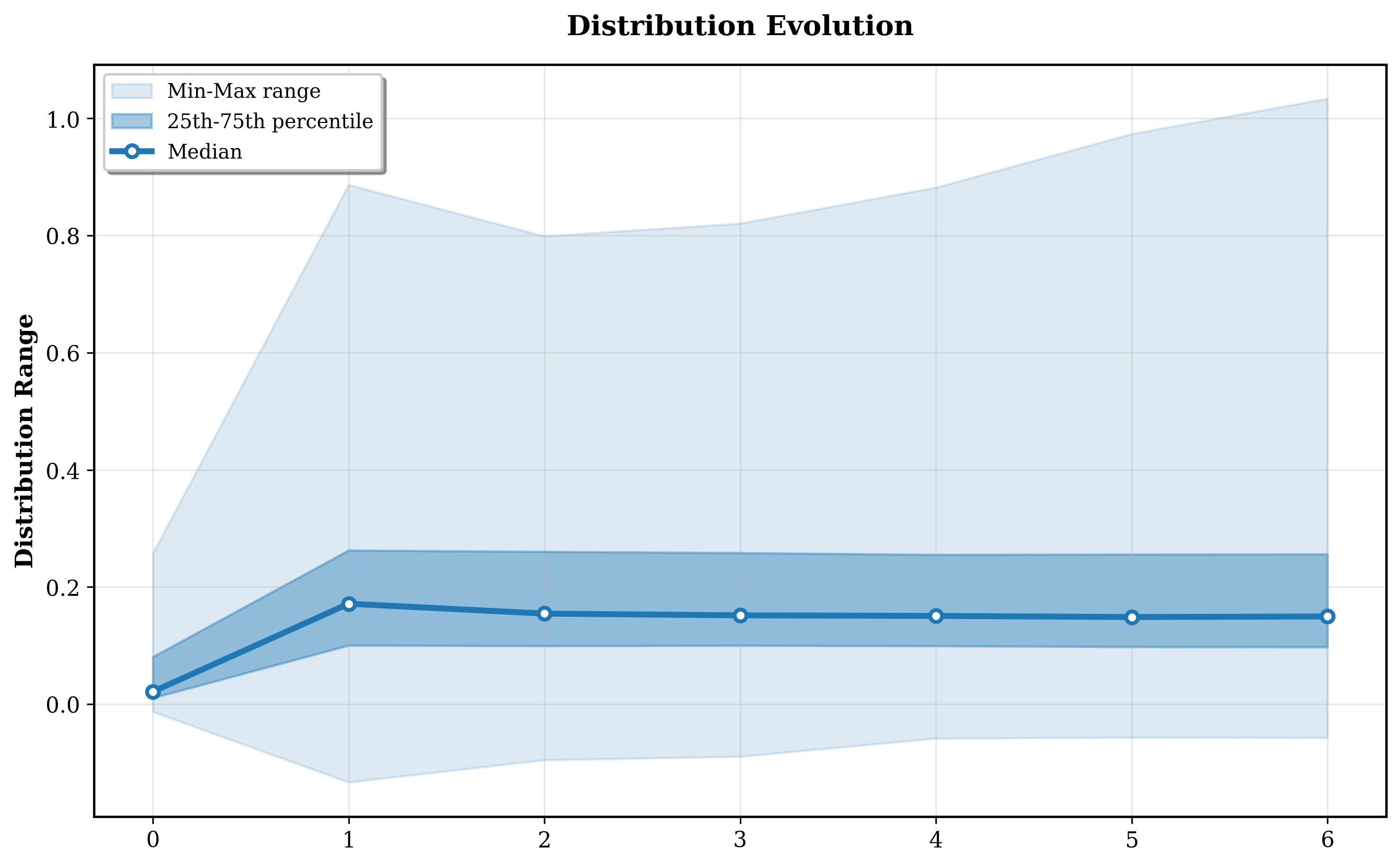}
  \caption{Redundancy of the model as measured by the spectral distribution of the Jacobian, starting with optimisation on spline space (iteration zero), followed by refinement in the projective spaces.}
  \label{param_redund}
\end{figure}

Investigating spectral distribution during fitting, we aggregated results from Section \ref{runav} in Figure \ref{param_redund}. The first iteration is the result of training within spline-space, and each iteration after are the projected space optimisation and refinement steps. There is a large increase in the variation and variability in the Jacobian range, confirming that the reduction in parameters has a significant impact on reducing redundancy and the impact of the 'nuisance space' mentioned earlier.

\subsubsection{Parameter Responses}\label{param_responses}

Training a network with varied hyperparameters on a [1,4,1] model with 5 iterations allowed on test functions introduced in Section \ref{ablation}, results were aggregated across functions.

\begin{figure}[H]
  \centering
  \includegraphics[width=0.75\linewidth]{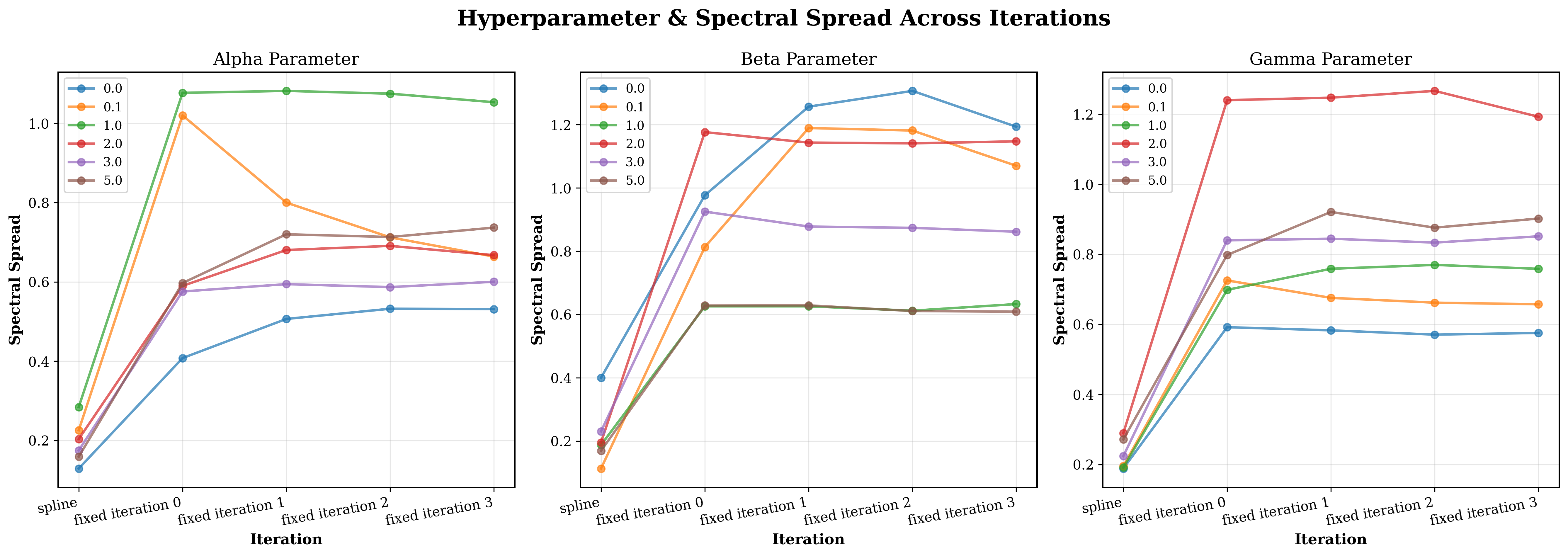}
  \caption{Parameter spread as measured by the standard deviation of the Jacobian values, across hyperparameter values. Each hyperparameter has a significant impact on the redundancy of the model, with $\alpha$ and $\gamma$ showing a clear maximum, while $\beta$ shows a marginal increase when set to zero on average.}
  \label{psa}
\end{figure}

Using standard deviation of Jacobian values (spectral spread) as a proxy for parameter redundancy, Figure \ref{psa} shows a clear impact of hyperparameter value on parameter redundancy, with maximisation of parameter impact occurring at (individually) $\alpha = 1.0$, $\beta = 0.0$ (marginally), and $\gamma = 2.0$. This presents a relationship that is not linear and ignores the highly interdependent nature of the hyperparameters in the optimisation process.

Also important is the prevalence of unfixed edges, which remain as splines. Results presented in Table \ref{hypothesis_test} show there is a higher than average number of unfixed edges when the $\beta$ value is smaller than $\gamma$, and when $\alpha + \gamma > \beta$. This follows the intuition that in these cases: minimising the number of edges over the representational compactness of the model results in twice the average rate of unfixed edges, since information that could be spread through multiple projective edges is compressed into fewer spline edges. Furthermore, when loss and edge number are both minimised more aggressively in comparison with the entropy term, the model is less likely to contain convergent edge representations.

\begin{table}[htbp]
  \centering
  \caption{Hypothesis Testing: Conditions Predicting Function Unfixing}
  \label{hypothesis_test}
  \begin{tabular}{l|c}
  \hline
  Condition & Mean Unfixed (\% overall edges) \\
  \hline
  $\beta < 0.5\gamma$ & 5.6 \\
  $\alpha + \gamma > \beta$ & 3.25 \\
  Baseline (all) & 2.6 \\
  \hline
  \end{tabular}
\end{table}

The reason these percentages are low is that the $R^{2}$ threshold is very low (0.6), since we don't mind over-simplification of edge in one iteration where we can revert if the regret is too high. This also explains why the hyperparameter $\beta$ appears counterintuitively to marginally discourage parameter importance when measured through the Jacobian's spectral values. The gamma term compensates by making edges redundant, reducing the likelihood of edge fixing. If the goal is to simplify the model into as few edges as possible, then by removing the entropy metric, the model could be optimised in spline space alone. However, this would not be a P-KAN model, and would not benefit from the reduced parameter space.

\subsection{Robustness Under Noise \& Architecture Variation}\label{runav}

Continuing experiments mentioned above, KAN and P-KAN architectures were tested under various noise conditions and with variations in model architecture in 1D and 2D. Gaussian, uniform, and salt-and-pepper noise were added to training data at SNR levels of 5, 10, 15, 20, and 30 dB, as well as a noise-free baseline. In terms of architecture, we measured minimal (neuron layers: [1,1]), tiny ([1,2,1]), small ([2,4,1]), and medium ([2,8,1]) sizes.

To standardise tests, the 2D heat equation was used as a benchmark for 2D experiments, and a simple wave equation was used for 1D experiments. Each experiment was repeated 5 times, and the average taken.

\begin{figure}[H]
    \centering
    \begin{subfigure}[b]{0.49\textwidth}
        \includegraphics[width=\textwidth]{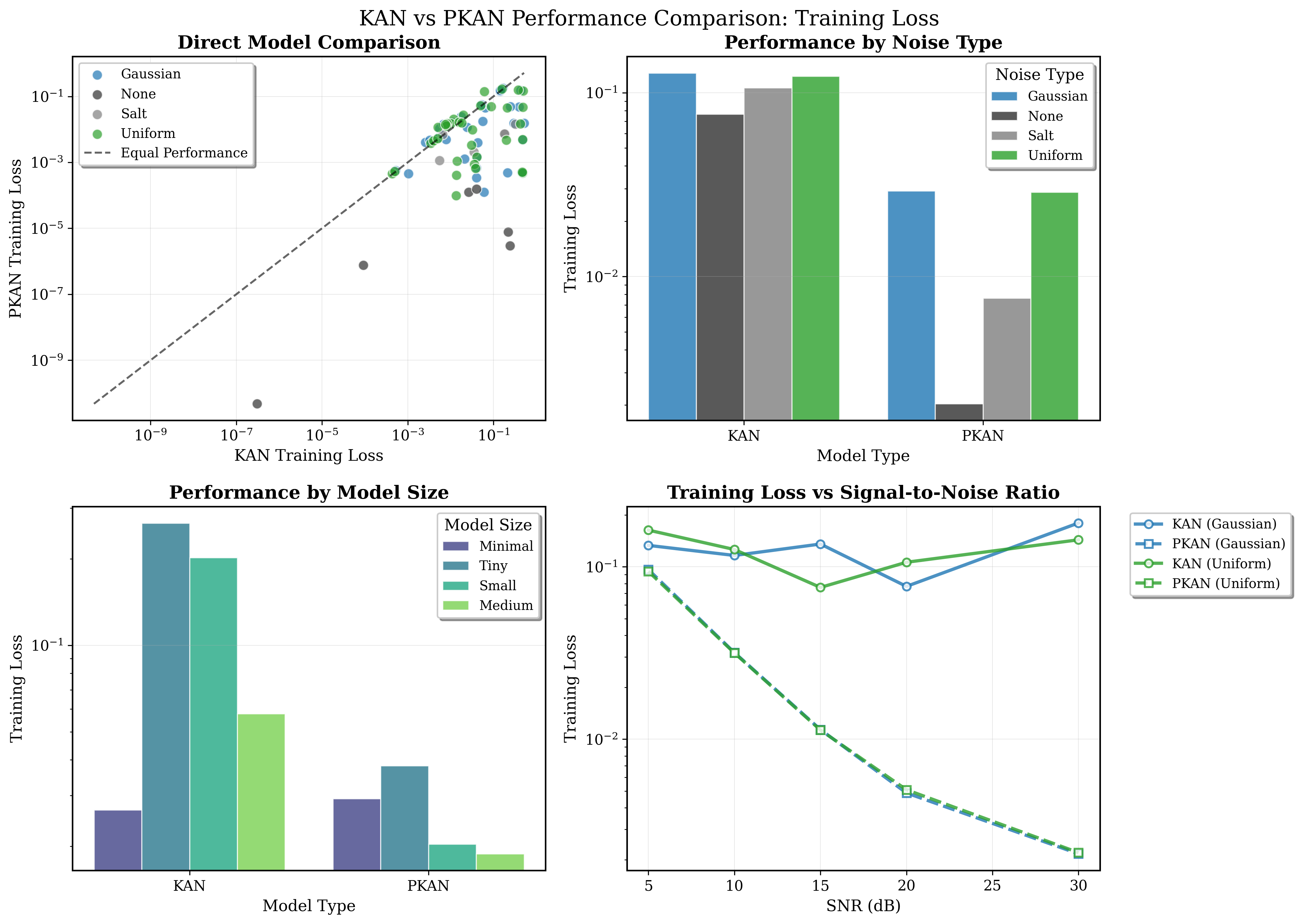}
        \caption{}
    \end{subfigure}
    \hfill
    \begin{subfigure}[b]{0.49\textwidth}
        \includegraphics[width=\textwidth]{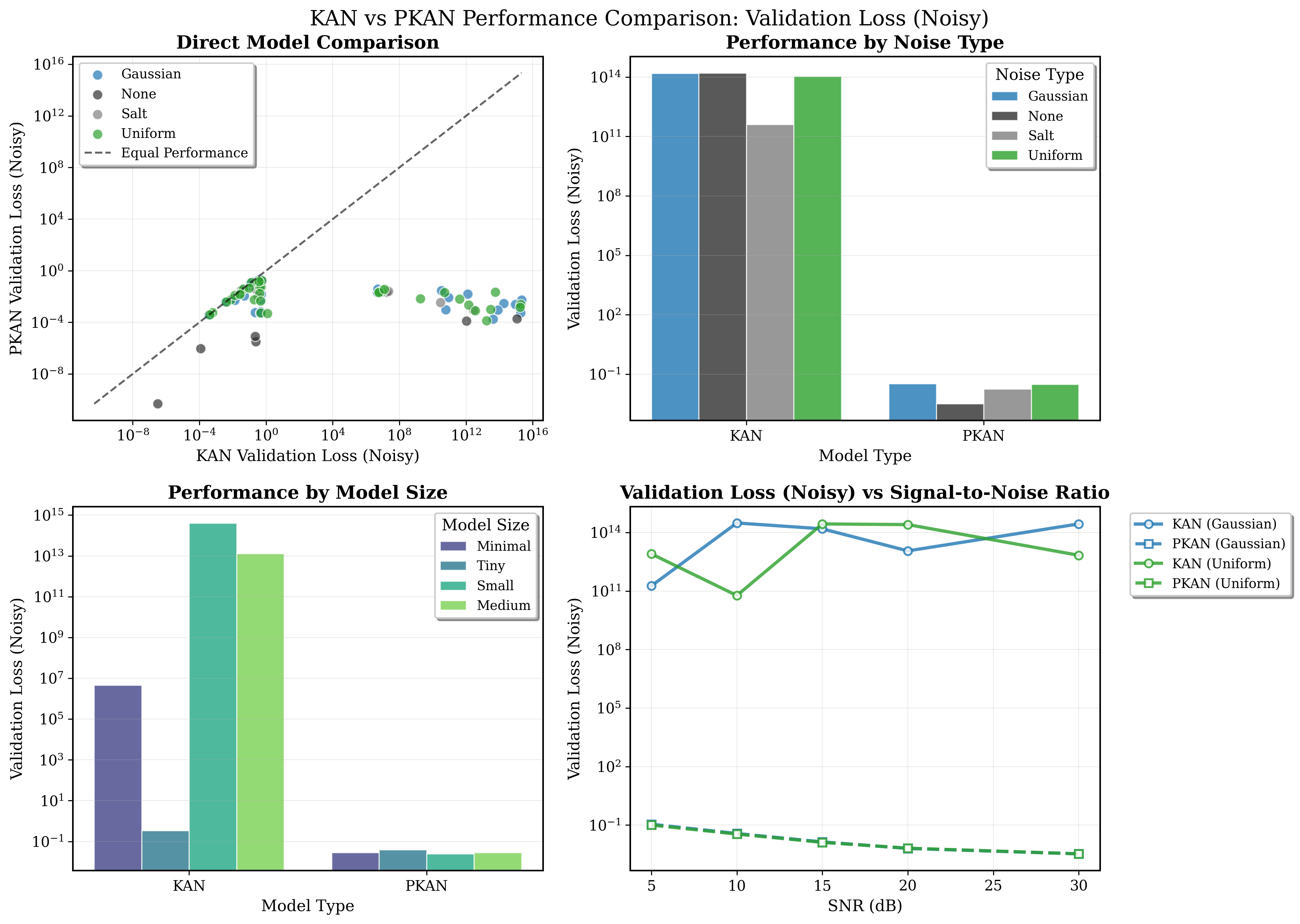}
        \caption{}
    \end{subfigure}
    \hfill
    \begin{subfigure}[b]{0.49\textwidth}
        \includegraphics[width=\textwidth]{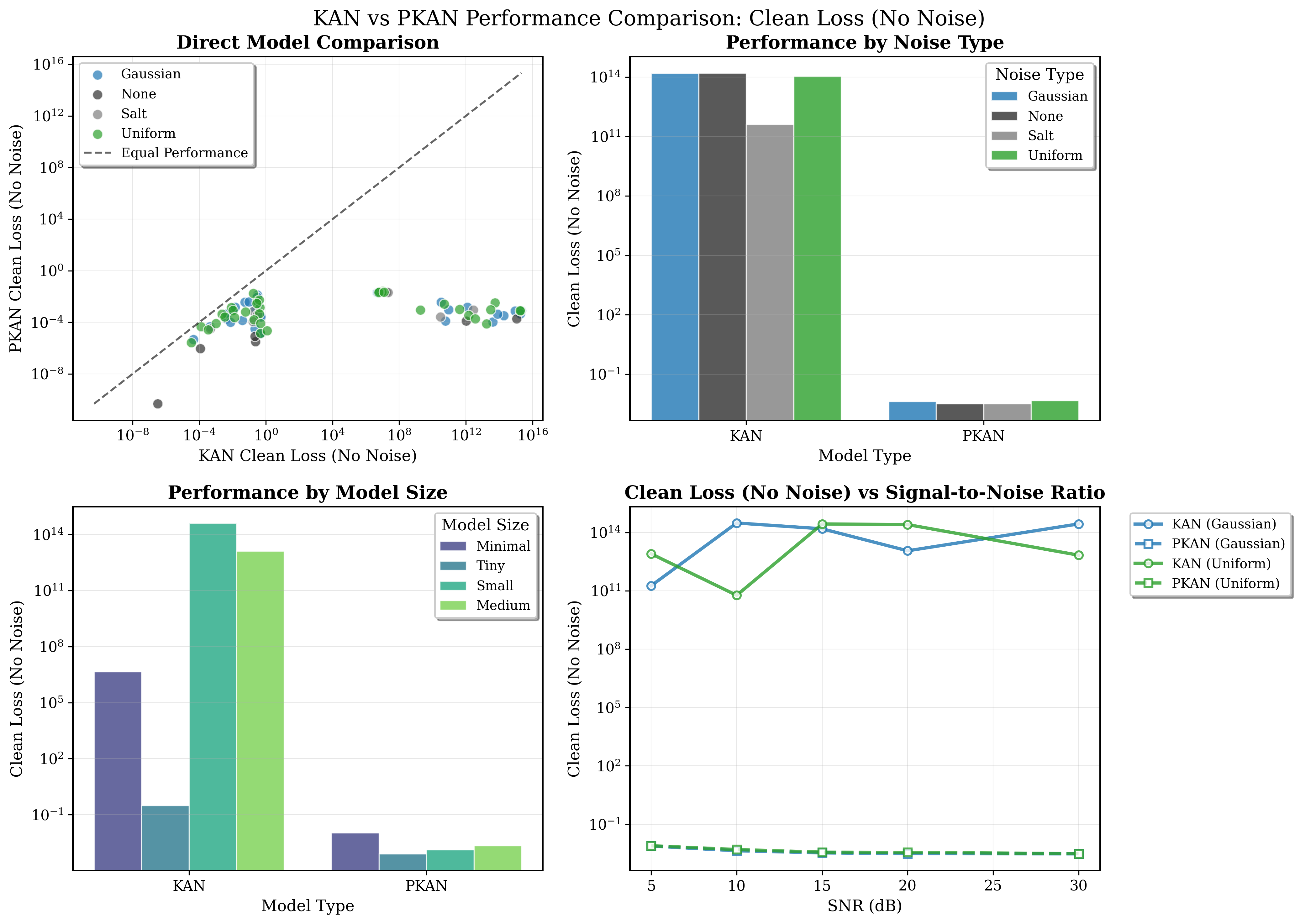}
        \caption{}
    \end{subfigure}
    \caption{The training loss graph, (a) shows that both model architectures converge well, largely, to the data given to them for all noise levels. However, when tested against the validation data - both noisy (b) and clean (c), the P-KAN approach does not over-fit to noise and handles random sampling rates significantly better than the KAN architecture, reflecting the earlier commentary on the nuisance space.}
    \label{loss_graphs}
\end{figure}

Figure \ref{loss_graphs} shows both models converge to training data relatively well, however both sampling and noise affect convergence of the KAN model far more than the P-KAN model. The added functional structure of the P-KAN model learns the data with far greater accuracy across scales, noise levels, and architectures.

\begin{figure}[H]
  \centering
  \includegraphics[width=0.75\linewidth]{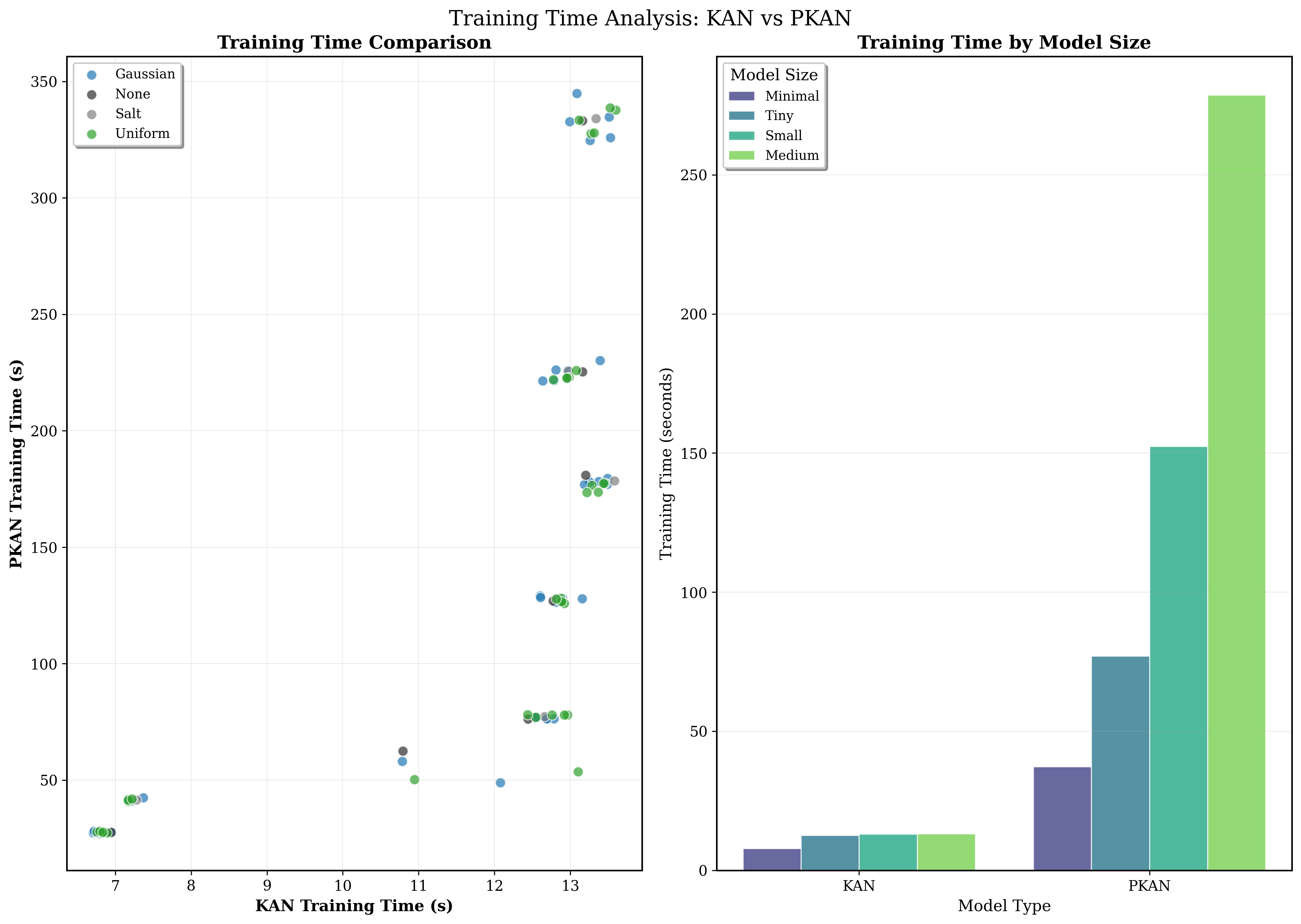}
  \caption{Average training time for P-KAN and KAN. KAN training scales well, while P-KAN is not as efficient with larger networks. We can see the training time is not correlated with noise type for P-KAN.}
  \label{training_time}
\end{figure}

Meanwhile, KAN outperforms on training time significantly, as shown in Figure \ref{training_time}, the P-KAN projection measurements and optimisation process providing a not insignificant computational cost.

However the stability in convergence towards learning the data's structure, seen when analysing both raw (noisy) validation loss and clean (no added noise) validation loss in Figure \ref{loss_graphs}, is significantly better in the P-KAN model across scales. An interesting feature of the KAN model is how it suffers under scale. While a single-edge network learns the input/output relationship well, added complexity does not lead to more optimal solutions. This implies there is a strong attraction to local minima introduced that become closer to the real minima the greater the size is. This is mirrored to an extent with the P-KAN model, however the minimal model is exceeded very quickly with additional neurons.

While the KAN model is capable of learning the data to within reason, it significantly overfits to the training data while the P-KAN model remains more generalisable across noise and size domains.

\subsection{Automated Fibre Placement}\label{afp}

\begin{figure}[H]
    \centering
    \begin{subfigure}[b]{0.49\textwidth}
        \includegraphics[width=\textwidth]{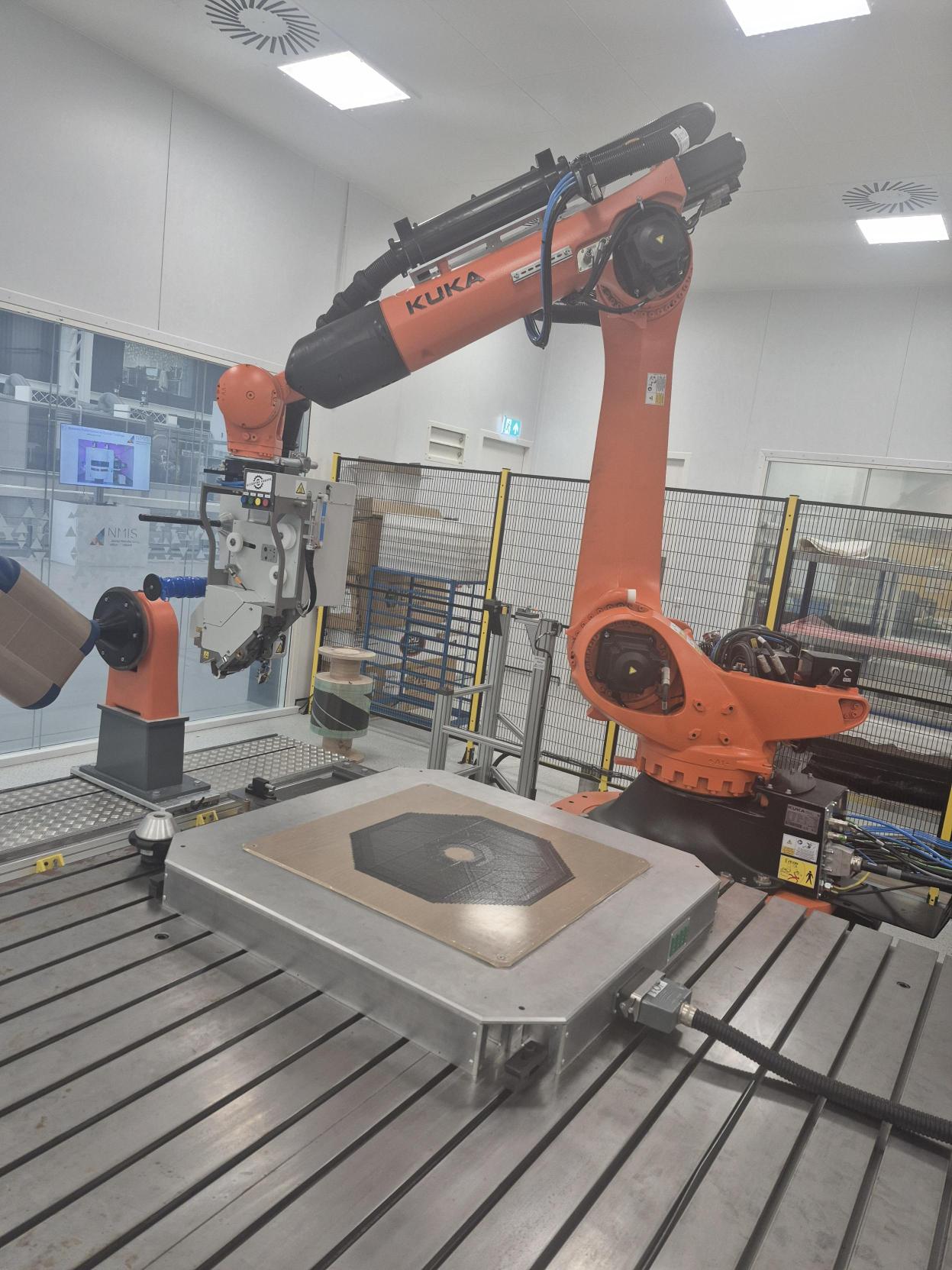}
        \caption{}
    \end{subfigure}
    \hfill
    \begin{subfigure}[b]{0.49\textwidth}
        \includegraphics[width=\textwidth]{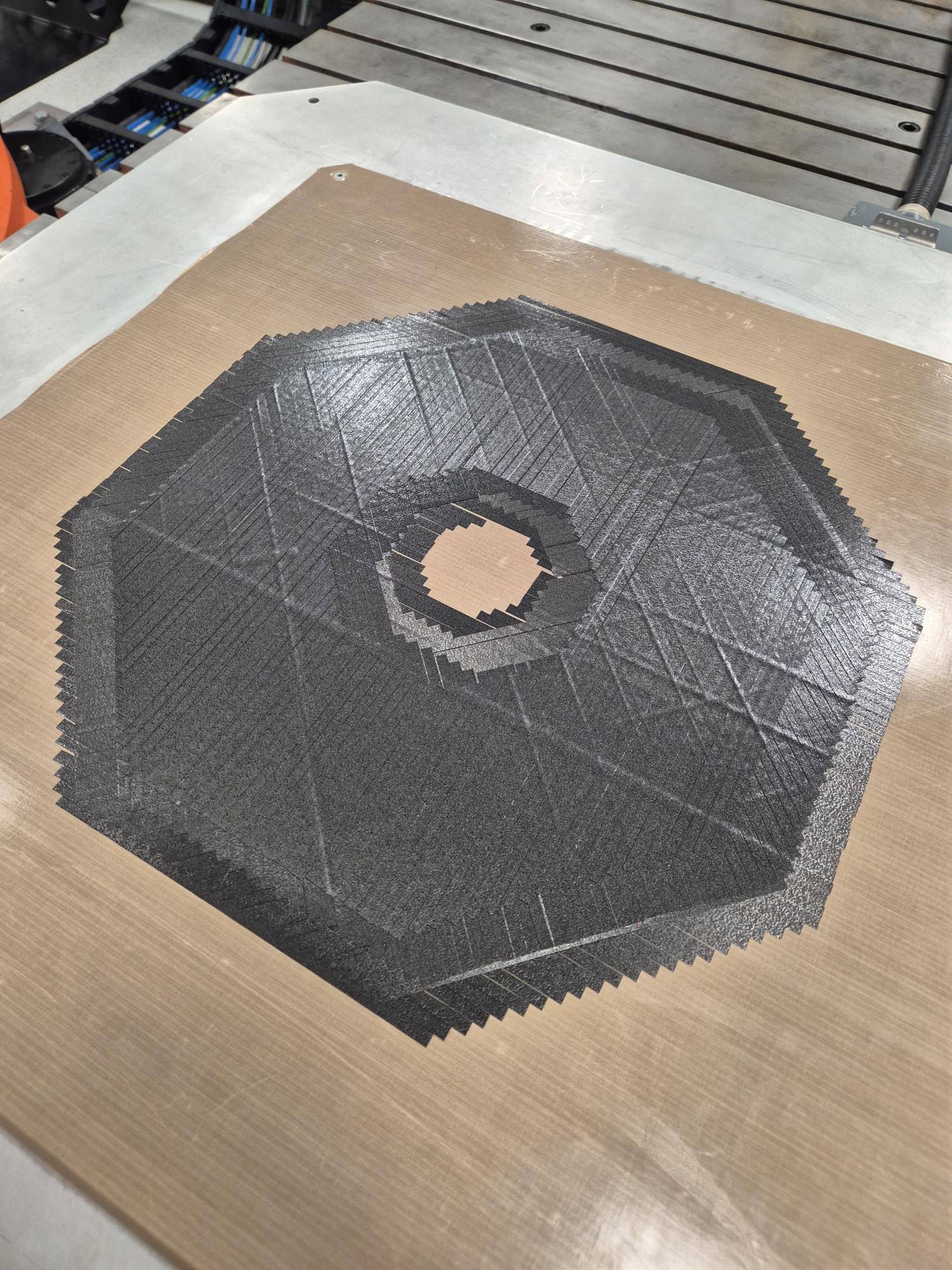}
        \caption{}
    \end{subfigure}
    \caption{The AFP cell, (a), involves a KR120 track-mounted robotic platform placing individual tape samples on a heated table, seen in the bottom right. The placement process is viewed through a ScanControl laser-line scanner parallel mounted on the tool head, collecting real-time data at 25Hz that is then passed on to an external control system. Cumulative tape placement creates designs such as in (b), an octogonal test-piece.}
    \label{afpcell}
\end{figure}

We seek to validate the process on industrially relevant data. An up-and-coming approach to composite manufacturing is Automated Fibre placement, laying tow-line strips of material onto a heated bed, later transported to an autoclave. While maintaining an edge over operator-handled composite samples in terms of layup speed and repeatability, there are several issues induced that require attention. Twisting of the samples, incorrect layup direction (wandering tow), and placement gaps are deemed critical defects, leading to scrappage after the completed layup process. A layup cell and example layup pattern, seen in Figure \ref{afpcell}, is equipped with a laser line scanner on the toolhead, measuring the tow lines during placement.

\begin{figure}[H]
    \centering
    \begin{subfigure}[b]{0.49\textwidth}
        \includegraphics[width=\textwidth]{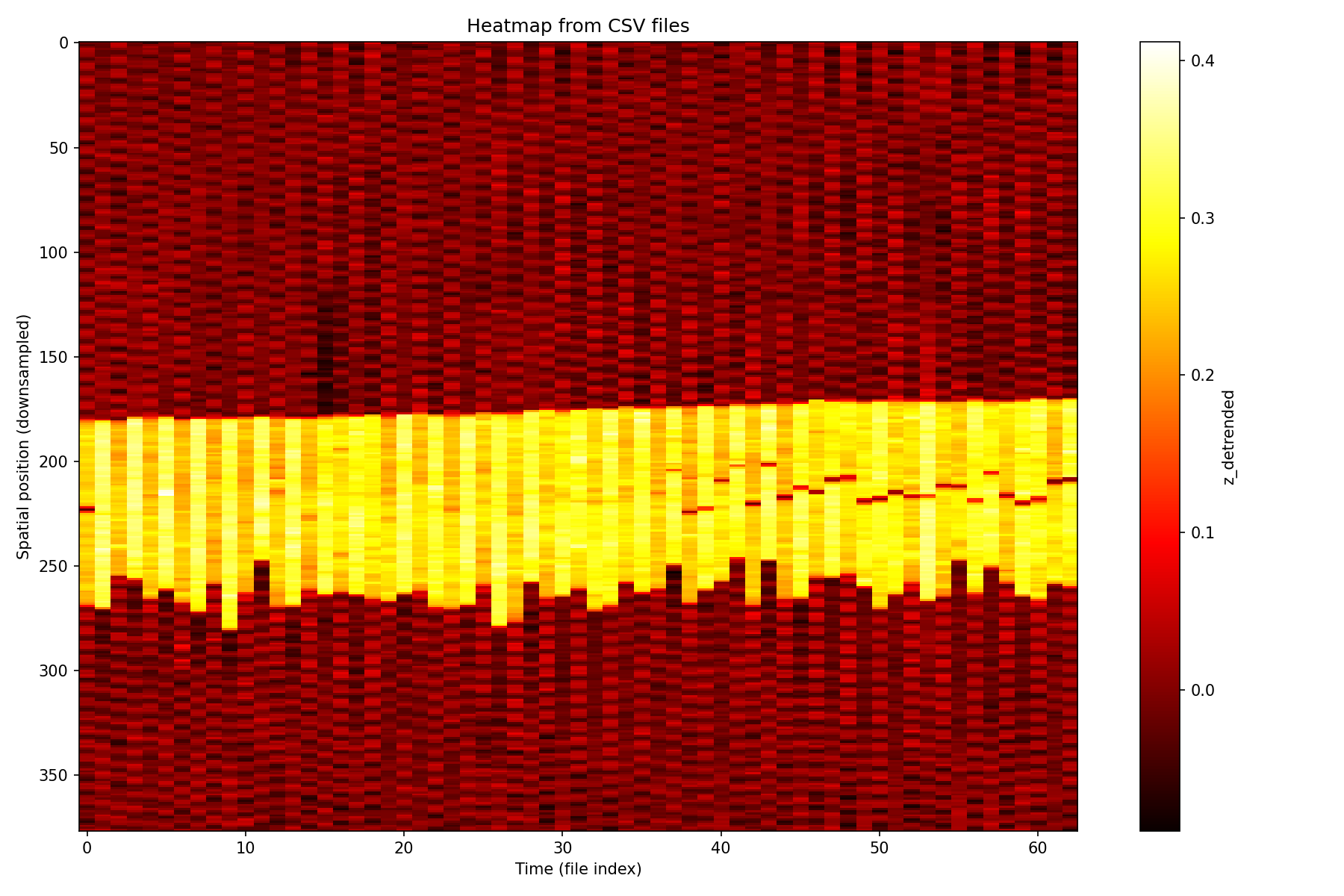}
        \caption{}
    \end{subfigure}
    \hfill
    \begin{subfigure}[b]{0.49\textwidth}
        \includegraphics[width=\textwidth]{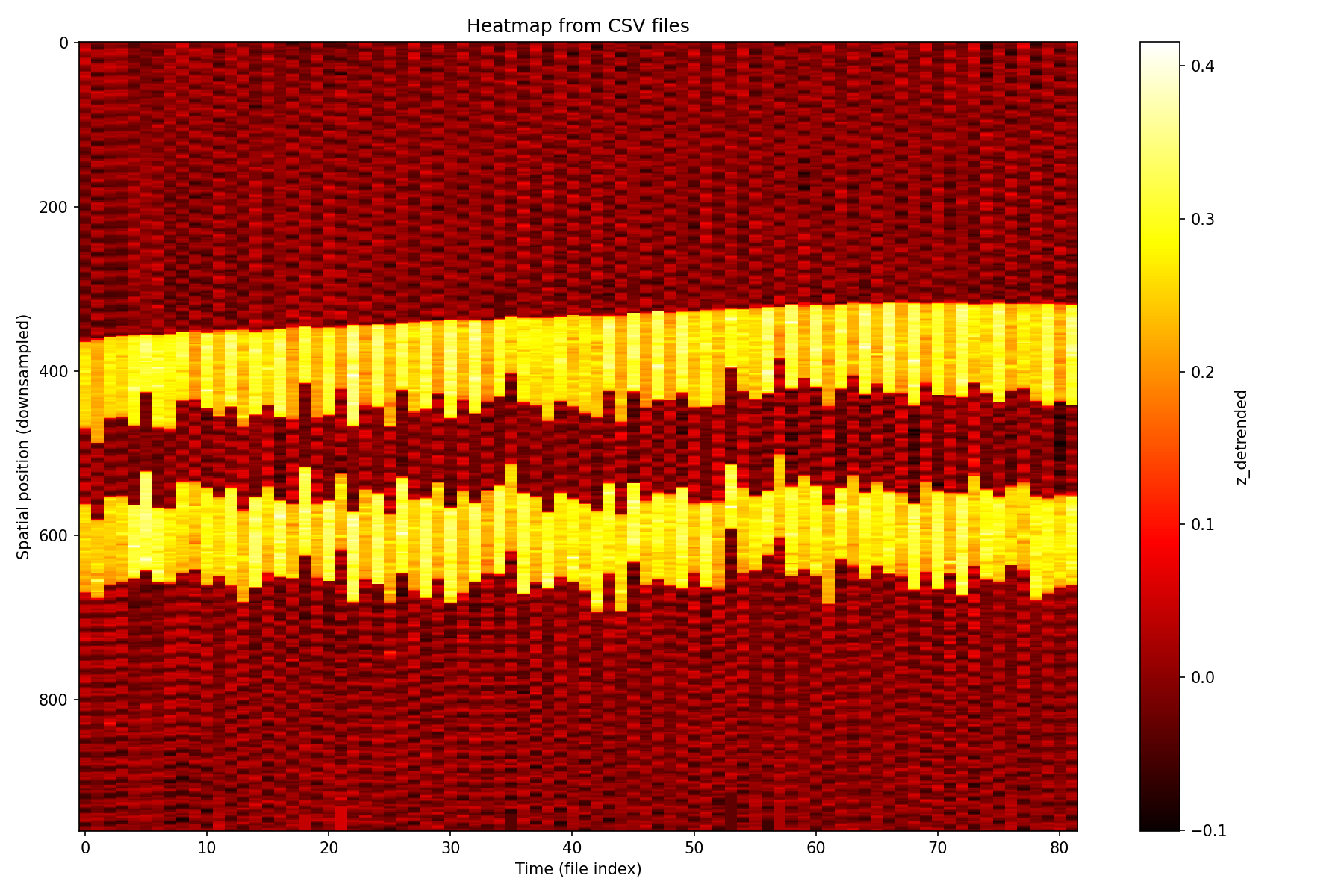}
        \caption{}
    \end{subfigure}
    \caption{Normalised laser-data map of tow segments. In (a), a single tow without defects is present, and in (b), two defect free parallel tows.}
    \label{laserlinescanner}
\end{figure}

Prior and ongoing research \cite{afpcite} has looked at online control patterns to ensure the tow lines are placed correctly, requiring high-fidelity laser data and an as-built vs as-designed discrepancy model. A predictive model capable of identifying the next stage of the tow's position, given the current laser scan would be of great benefit to this process. Due to the reflective properties of the composite tow, the data is highly noisy, with significant outliers as can be seen for samples in Figure \ref{laserlinescanner}.

\subsubsection{Model \& Training}

\begin{figure}[H]
    \centering
    \includegraphics[width=\textwidth]{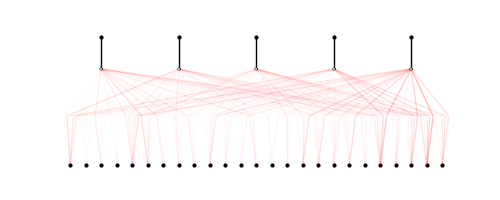}
    \caption{The network architecture, with $5\times5$ collected laser-data inputs flattened out, and the next laser-line-scan's adjacent data prediction as the outputs.}
    \label{model}
\end{figure}

The data collected consisted of laser line scans aligned such that individual line-scan data is in y, and time-index in x. We trained a $25\rightarrow 5$ network to act as a CNN, taking a 5x5 grid of pixel values, and predicting the 5 adjacent laser data points for the next scan, the network and diagram of the in/out data seen in Figure \ref{model}.

The training data was collected from 5 separate tow placements. Material costs were sought to be minimised, and only 14 samples taken to train the network, a full accounting of which is in Table \ref{data_collected}.

\begin{table}[htbp]
\centering
\caption{Data collected for AFP model training. Since we need a $5\times 5$ input data stream, the useable time steps is reduced by 5 from the total time-steps.}
\label{data_collected}
\begin{tabular}{l|c|c}
\hline
Type & Number of Tows & Average Tow Length (time-steps) \\
\hline
Single-tow & 6 & 41.5\\
Parallel-tow & 4 & 63.25 \\
Overlapped-tow & 4 & 82 \\
\hline
\end{tabular}
\end{table}

\subsubsection{Results}

\begin{figure}[H]
    \centering
    \begin{subfigure}[b]{0.49\textwidth}
        \includegraphics[width=\textwidth]{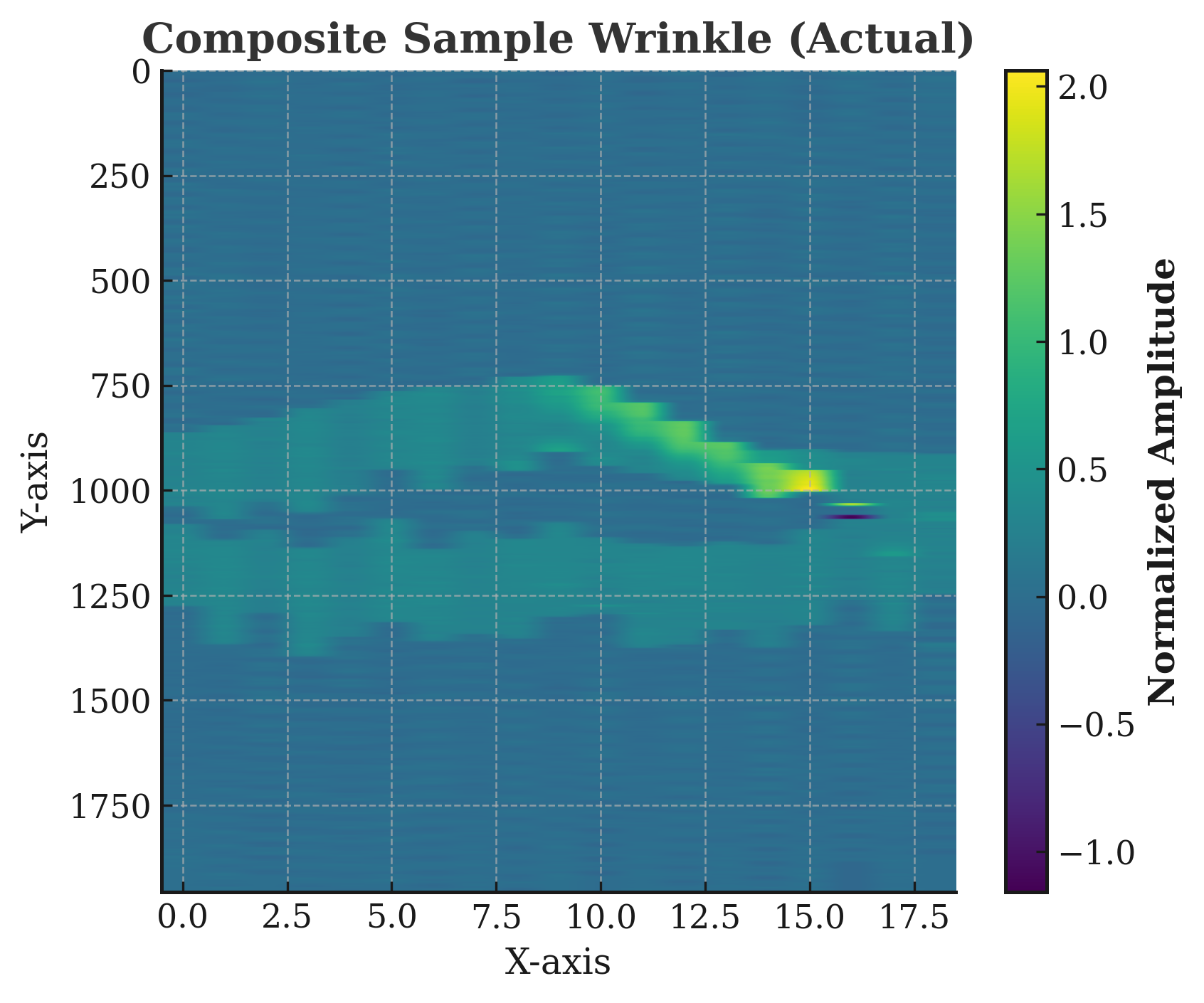}
        \caption{}
    \end{subfigure}
    \hfill
    \begin{subfigure}[b]{0.49\textwidth}
        \includegraphics[width=\textwidth]{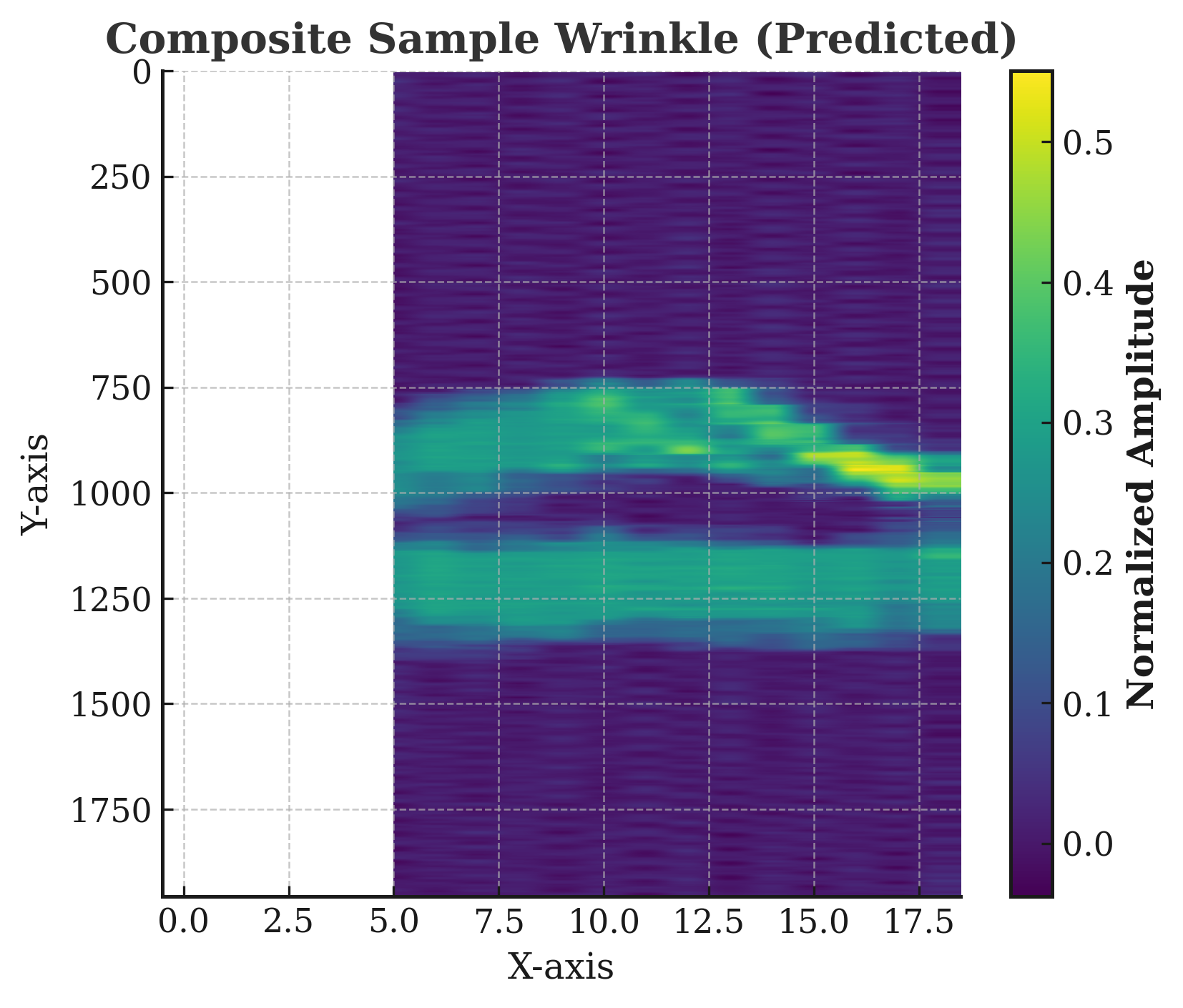}
        \caption{}
    \end{subfigure}
    \caption{Two parallel composite tows, with one ending in a wrinkle, original data in (a), and next-step-predicted in (b). The model has succeeded in predicting the next step for an unseen sample, up to scaling.}
    \label{wrinkle_predict}
\end{figure}

Even without training on defective samples, the model is able to accurately predict the tow's placement, as shown with an artificial wrinkled sample in Figure \ref{wrinkle_predict}. The model, though shallow, has learned the shape of the input data, and more importantly can extrapolate to new different data heights. While this is a simple example, it is a non-trivial example given the raw data used. The closest approach to this is \cite{KOPTELOV2025112655}, using ideal data in a static rig, implemented with a deep neural architecture equipped with LSTM nodes. In comparison, we have obtained accurate predictions with a single-layer network that is noise resilient and can extrapolate well. While graphically identical, we can see the relative scales are different. We therefore normalise the data to combat this in the rest of this section.

For the octogonal test-piece seen in Figure \ref{afpcell}, the model was able to predict the next laser-line-scan's adjacent data points with a mode RMSE score of $\sim 0.08$.

\begin{figure}[H]
    \centering
    \includegraphics[width=\textwidth]{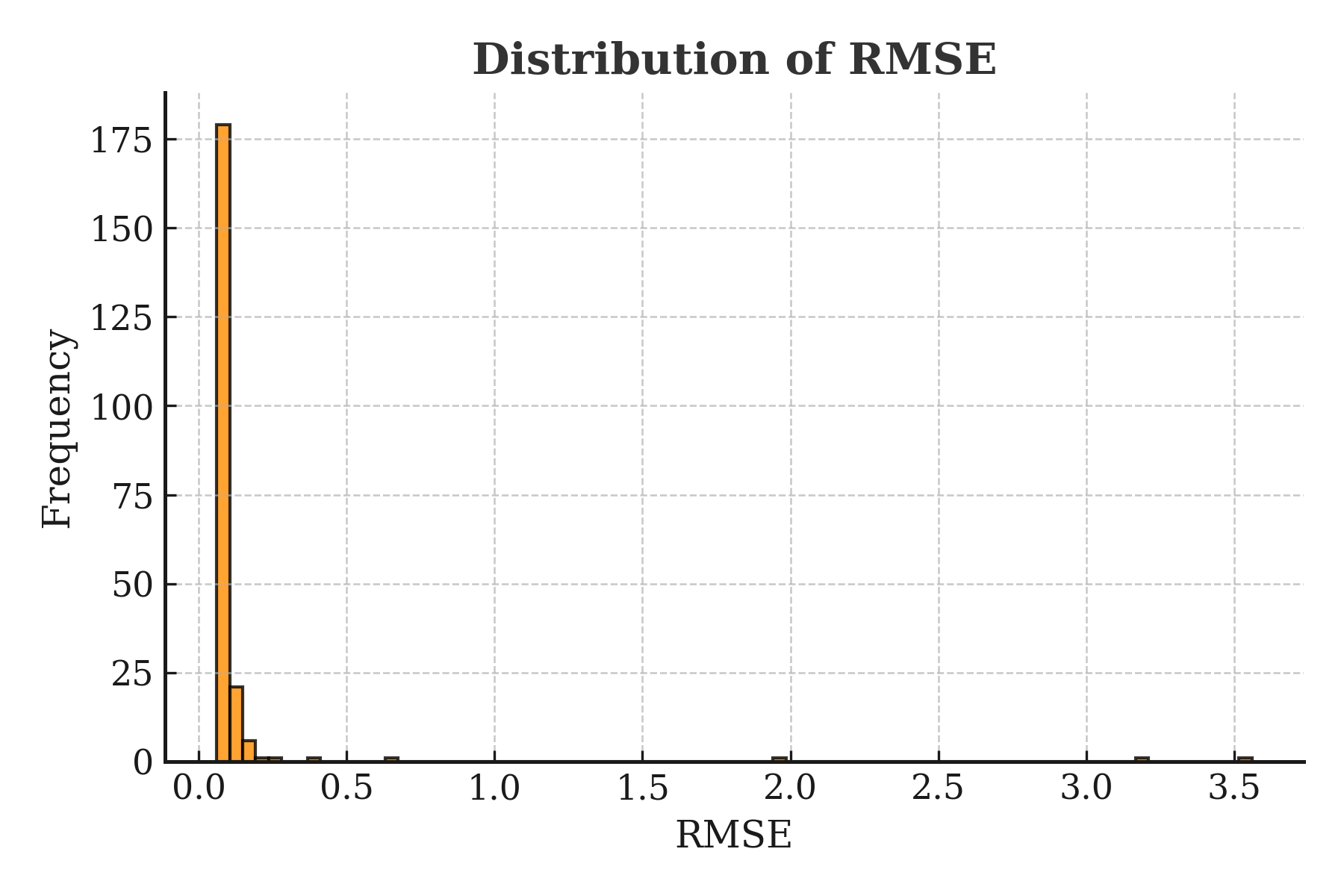}
    \caption{RMSE histogram of the model's predictions on the octogonal test-piece, with a mode of $\sim 0.08$}
    \label{rmseerr}
\end{figure}

\begin{figure}[H]
    \centering
    \begin{subfigure}[b]{0.49\textwidth}
        \includegraphics[width=\textwidth]{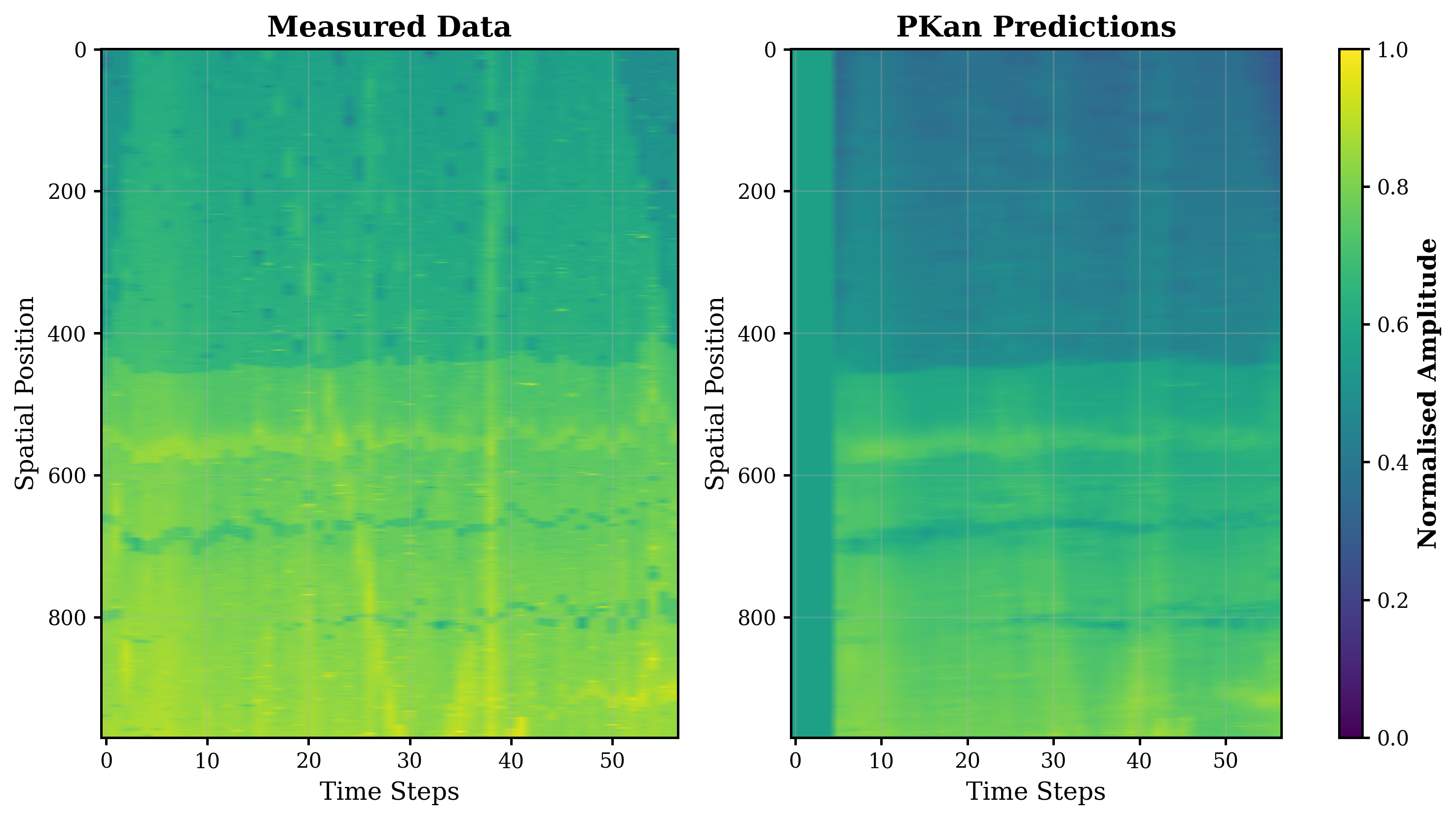}
        \caption{}
    \end{subfigure}
    \hfill
    \begin{subfigure}[b]{0.49\textwidth}
        \includegraphics[width=\textwidth]{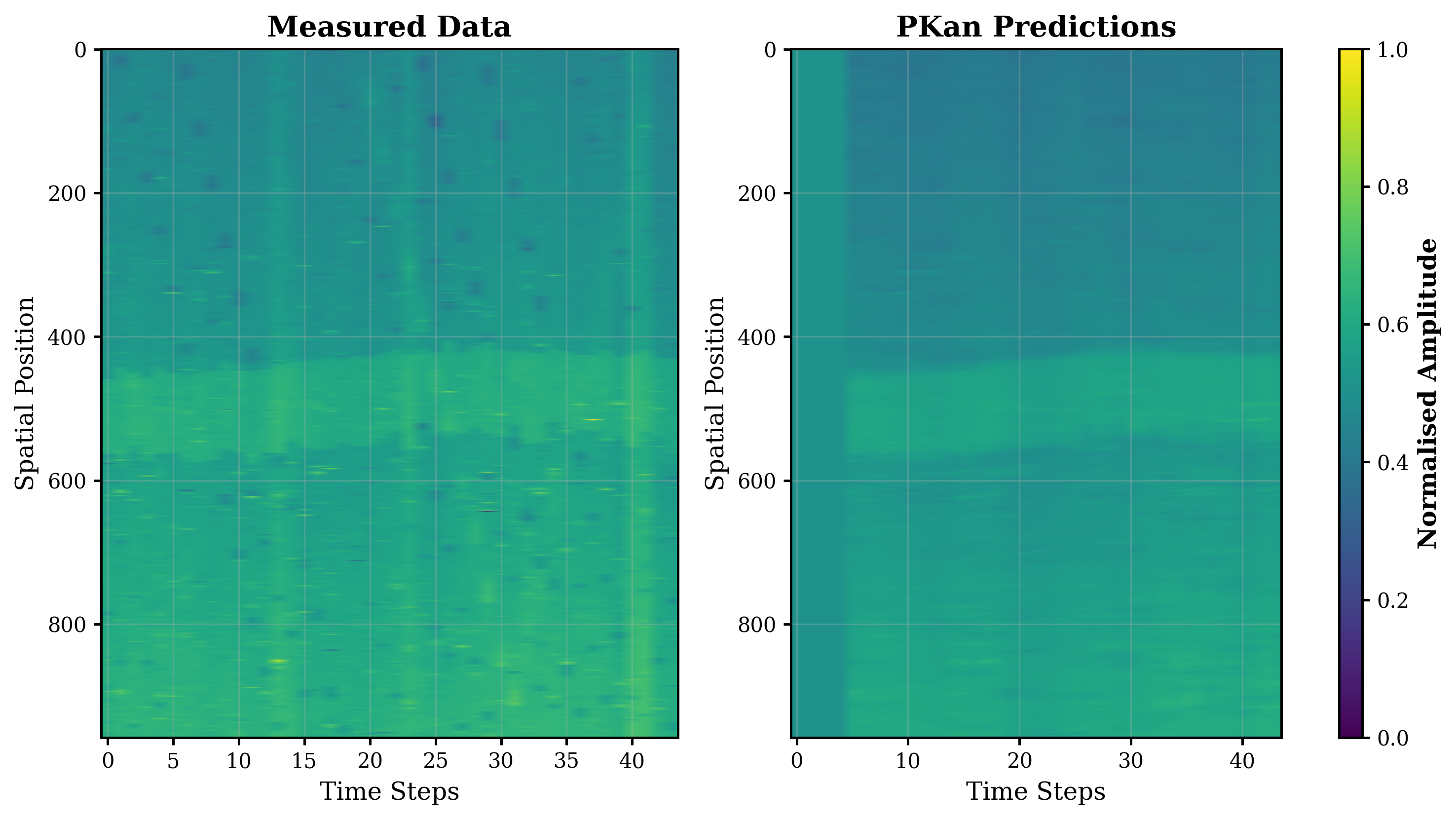}
        \caption{}
    \end{subfigure}
    \begin{subfigure}[b]{0.49\textwidth}
        \includegraphics[width=\textwidth]{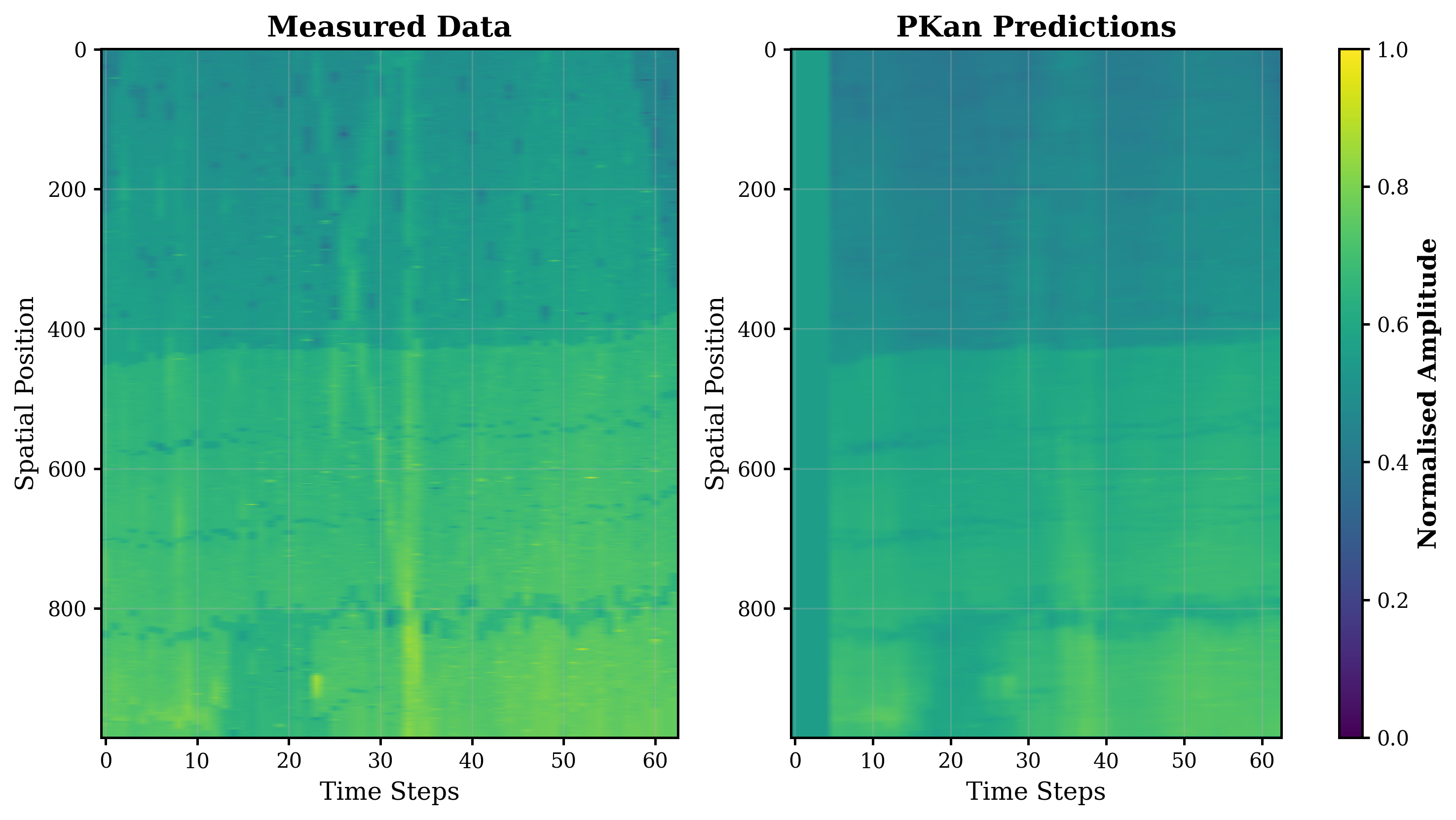}
        \caption{}
    \end{subfigure}
    \hfill
    \begin{subfigure}[b]{0.49\textwidth}
        \includegraphics[width=\textwidth]{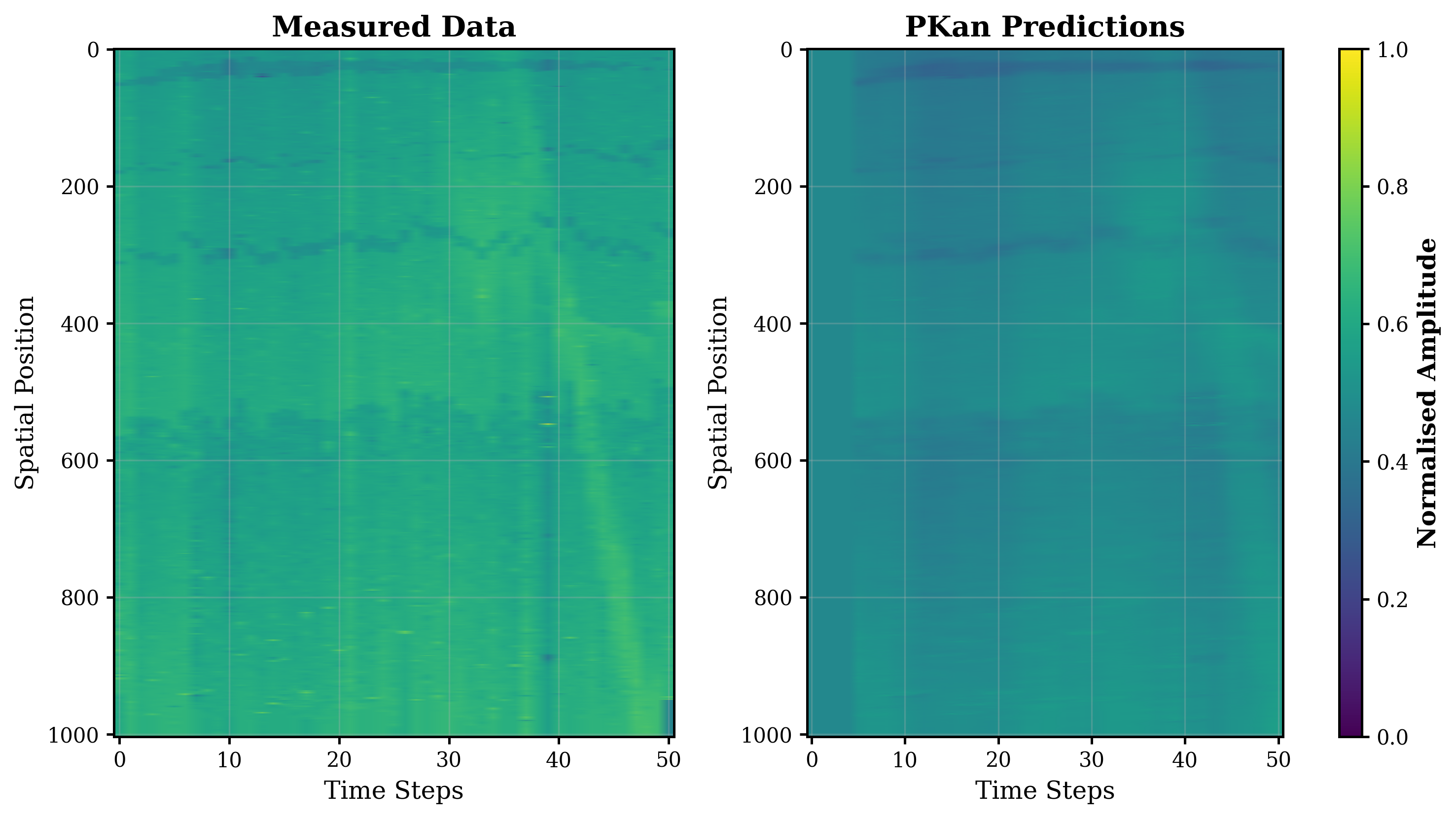}
        \caption{}
    \end{subfigure}
    \caption{Prediction results for: (a,b,c) the fourth layer of the AFP layup, (d) the third layer. In (c), we can see the central cut-out of the octogonal layup in \ref{afpcell}.}
    \label{afp_awesome}
\end{figure}

Some results for the layup process are shown in Figure \ref{afp_awesome}. Despite the relatively low-quantity and quality training data, with no prior filtering of the data the model has learned a predictive model that is relatively smooth compared to actual data, demonstrating its ability to extract structure and ignore structure-less components, enabling a clean analysis for a predictive maintenance framework. And although the training data was collected on a heated bed with no pre-existing reflective composite substrate, it has been able to generalise to this exact scenario, as seen with the 4th layer predictive results.

\section{Critical Analysis}\label{critical_analysis}

\subsection{Summary of Key Findings}

The experimental results demonstrate P-KAN's performance across three distinct evaluation domains. In systematic ablation studies, P-KANs achieved $R^2 > 0.85$ across the majority of the tested model complexities whilst reducing parameters by over 80\% (from 23 to 4 degrees of freedom per edge). The entropy-driven selection mechanism successfully identified appropriate functional spaces without \textit{a priori} specification, though no single space dominated universally (Figure \ref{func_abl}).

Under adverse conditions, P-KANs maintained validation losses of $<\mathcal{O}(10^{-1})$ across all noise levels (5-30 dB SNR), whilst standard KANs exhibited losses exceeding $10^{13}$ under identical conditions. The spectral analysis confirmed a 340\% increase in Jacobian variance after functional projection, quantifying the reduction in parameter redundancy.

Industrial deployment on AFP manufacturing validated practical utility: with only 14 training samples, the model achieved RMSE $\approx 0.08$ on complex geometries and successfully generalised to multi-layer configurations without retraining.
The discovered dominant functional form of Fourier series aligns with classical image reconstruction techniques.

\subsection{Theoretical Implications}

The parameter reduction mechanism addresses the nuisance space problem more effectively than anticipated. The increase in Jacobian spectral spread (Figure \ref{param_redund}) confirms that functional projection concentrates parameter influence, reducing the dimensionality of the effective optimisation space.
This concentration appears to prevent the training instabilities observed in standard KANs at higher model complexities.

The phase transition behaviour observed at $\beta < 0.5\gamma$ (Table \ref{hypothesis_test}) suggests the existence of critical thresholds in the optimisation landscape.
Below this threshold, entropy minimisation cannot overcome edge regularisation, minimising the network's size at the detriment of function-space discovery.
This finding has implications for hyperparameter scheduling: initial training could use high $\beta$ values to encourage exploration, followed by reduction to allow convergence.
Further, is the interest is in network simplification and training time, then the entropic terms could be removed entirely.

The mixed functional space results indicate that complex systems naturally decompose into heterogeneous functional representations. Different edges converging to different spaces suggests that networks can discover multi-physics representations where distinct physical processes—each governed by different mathematical structures—interact within a single model.

\subsection{Practical Advantages}

The noise resilience demonstrated across multiple conditions appears to stem from functional space constraints acting as implicit regularisation.
By restricting edge functions to spaces spanned by known mathematical structures, P-KANs encode smoothness priors that align with physical processes. This explains the successful extraction of signal from noisy data in both synthetic and industrial tests.

Data efficiency represents perhaps the most significant practical advantage.
The AFP application's success with 14 training samples—compared to typical deep learning requirements of thousands—makes P-KANs viable for domains where data collection is expensive or limited. The ability to generalise from single lay-up to multiple lay-up structures without retraining suggests the framework captures fundamental relationships rather than memorising patterns.

The automatic discovery of interpretable functional forms provides value beyond prediction accuracy.
In the AFP case, identifying Fourier series as the dominant representation aligns with established image processing techniques, providing confidence in the model's validity. This interpretability could facilitate regulatory approval in safety-critical applications.

\subsection{Limitations and Trade-offs}

Computational overhead remains the primary limitation. The $12.4\times$ increase in training time stems from unavoidable projection operations across multiple functional spaces. Whilst P-KANs typically converge in fewer epochs, partially offsetting per-iteration costs, this overhead limits applicability to real-time or online learning scenarios.

The current restriction to classical orthonormal bases may prove insufficient for exotic phenomena.
Turbulent flows, fractal patterns, and anomalous diffusion processes may require functional spaces not included in the current implementation.
The framework's extensibility to arbitrary spaces remains theoretically possible but computationally challenging.

Amplitude scaling issues, observed in both synthetic tests and AFP predictions, suggest that the separation of shape from scale inherent in orthonormal projections may limit quantitative accuracy.
Applications requiring precise magnitude predictions may need additional calibration or hybrid approaches.

The hyperparameter sensitivity revealed in Figure \ref{psa} complicates deployment.
The non-monotonic relationships and interdependencies between $\alpha$, $\beta$, and $\gamma$ lack intuitive interpretation, requiring empirical tuning for each application domain.
Adaptive or learned hyperparameter schedules could address this but would further increase computational requirements.

\subsection{Scalability Considerations}

The projection computation time is observed to scale linearly with the number of edges, implying that very large networks may become impractical in comparison to the KAN architecture, whose training time scales more favourably.
However, the P-KAN's ability to perform well across scales is notably better than KAN, which suffers from local minima issues as complexity increases. This suggests a trade-off: for small to medium-sized networks where accuracy and robustness are paramount, P-KANs are advantageous; for networks whose only concern is training speed, KANs may remain preferable.

\subsection{Implications for Future Development}

The success with current functional spaces motivates extension beyond functional inner-product spaces spaces to more general mathematical structures.
Functions satisfying differential operators that don't form traditional vector spaces—such as solutions to nonlinear or fractional differential equations—could provide more natural representations for certain phenomena. The entropy-based selection mechanism could potentially extend to these cases through appropriate distance metrics on solution manifolds.

The observed tendency for different edges to select different functional spaces suggests that mixed-operator representations might emerge naturally in complex systems.
This points towards a unified framework where both the functional forms and their governing equations become learnable components, potentially bridging P-KANs with recent advances in neural differential equations and physics-informed neural operators.

The industrial success despite minimal data and significant noise validates the practical value of functional constraints.
However, the computational trade-offs suggest that hybrid approaches—applying P-KAN constraints selectively based on noise levels or data availability—may prove optimal for production deployment. Areas with abundant clean data could use standard approaches for speed, whilst critical components with limited noisy data employ P-KAN's robustness.

\section{Conclusions}\label{conclusions}

This work introduced Projective Kolmogorov-Arnold Networks (P-KANs), addressing fundamental inefficiencies in KAN architectures through entropy-driven functional space projection. The framework demonstrates that structured functional representations can simultaneously improve parameter efficiency, noise robustness, and model interpretability whilst maintaining the universal approximation guarantees of standard KANs.

\subsection{Summary of Contributions}

The primary theoretical contribution lies in the unified treatment of functional vector spaces within a single optimisation framework. By computing projection coefficients across multiple candidate spaces and using entropy as a selection metric, P-KANs automatically discover appropriate representations without \textit{a priori} specification. The regret-based adaptive mechanism ensures robustness when initial selections prove suboptimal, enabling recovery from poor local minima.

Experimentally, P-KANs achieved an 83\% reduction in parameters (from 23 to 4 degrees of freedom per edge) whilst maintaining $R^2 > 0.85$ for more than $60\%$ of cases across test functions. Under noise conditions causing standard KAN divergence to losses exceeding $10^{13}$, P-KANs maintained $<\mathcal{-1}(10^{-3})$ losses, outperforming every KAN when tested with validation data. The industrial AFP application validated practical utility, achieving RMSE $\approx 0.08$ with only 14 training samples and successfully generalising to unseen multi-layer configurations.

\subsection{Future Directions: Structured Functional Spaces}

The current framework's foundation on functional vector spaces with well-defined inner products represents both a strength—enabling computational tractability—and a limitation. Many physical phenomena are more naturally expressed as solutions to differential operators, removing the need for complex projection calculations.

Consider extending the framework to edge functions satisfying general differential equations:
\begin{equation}
    \mathcal{L}[f] = \lambda f
\end{equation}
where $\mathcal{L}$ represents differential operators beyond the harmonic equations underlying current bases.
Imposition of such structures could enable efficient and generalisable representations that enforce representational coherence, and learn structure with minimal prior knowledge of the data's content.

\subsection{Practical Implications}

The demonstrated data efficiency: achieving industrial-grade performance from minimal training samples positions P-KANs for domains where data acquisition is expensive or limited. Applications in materials characterisation, medical diagnostics, and quality control, where each measurement incurs significant cost, could particularly benefit from this efficiency.
Further, the randomised subsampling reflects a real-world scenario where data is often irregularly spaced and noisy. Traditional models require extensive cleaning by highly trained operators, a current bottleneck in actual model development. This is reflected in the poor validation data for KAN models, whereas the P-KAN model achieves consistently high accuracy despite the noise and random sparsity.

The interpretability of discovered functional forms provides value beyond prediction accuracy. In safety-critical applications, understanding model decisions through explicit mathematical relationships enables validation against domain knowledge and facilitates certification processes.

Hybrid deployment strategies could balance computational costs against robustness requirements. Critical components requiring high reliability under noisy conditions could employ P-KAN constraints, whilst less sensitive components use standard approaches for computational efficiency.

\subsection{Closing Remarks}

P-KANs represent a step towards neural architectures that discover rather than impose mathematical structure. By demonstrating that entropy-driven functional projection can improve multiple aspects of network performance simultaneously, this work suggests that further structuring of the functional space beyond traditional vector spaces could yield additional benefits.

The transition from viewing neural networks as universal approximators to understanding them as discoverers of mathematical structure opens new research directions. As we extend beyond vector spaces to general differential operators, we may find that the most effective neural architectures are those that can identify and exploit the natural mathematical structures inherent in the phenomena they model.

Successful application of P-KANs within an industrial AFP application, using minimal data and encountering significant noise highlights the po. However, the true potential lies not in the specific functional spaces employed here, but in the framework's extensibility to more general mathematical structures. This extensibility, combined with the demonstrated benefits, motivates continued investigation into structured functional representations for interpretable and efficient machine learning.

However, there is a lot of work necessary in enhancing the scope of the process described. The analysis presented is limited to known functional vector space implementations with known GPU-enhanced codebases.
It is the goal of the authors to extend this work by investigating imposition of more generalised structures within the functional spaces from different perspectives that may reduce overal computation time, implementing solutions with optimised software components (for example, integrated with ROS2 through C++/CUDA), and investigating derived structural similarities between datasets.

\section*{Acknowledgments}
The authors would like to thank the Lightweight Manufacturing Centre (LMC) for manufacturing the test piece in order to provide the industrial data for this study.
We would further like to thank the Applied Industrial Intelligence ($\text{AI}^{2}$) centre, and Department for Design, Manufacturing and Engineering Management (DMEM) for their support during the production of this paper.

\newpage
\bibliographystyle{unsrtnat}  
\bibliography{bibliography}   

\begin{thebibliography}{25}
\providecommand{\natexlab}[1]{#1}
\providecommand{\url}[1]{\texttt{#1}}
\expandafter\ifx\csname urlstyle\endcsname\relax
  \providecommand{\doi}[1]{doi: #1}\else
  \providecommand{\doi}{doi: \begingroup \urlstyle{rm}\Url}\fi

\bibitem[Liu et~al.(2025)Liu, Wang, Vaidya, Ruehle, Halverson, Soljačić, Hou,
  and Tegmark]{liu2025kankolmogorovarnoldnetworks}
Ziming Liu, Yixuan Wang, Sachin Vaidya, Fabian Ruehle, James Halverson, Marin
  Soljačić, Thomas~Y. Hou, and Max Tegmark.
\newblock Kan: Kolmogorov-arnold networks, 2025.
\newblock URL \url{https://arxiv.org/abs/2404.19756}.

\bibitem[Bozorgasl and
  Chen(2024)]{bozorgasl2024wavkanwaveletkolmogorovarnoldnetworks}
Zavareh Bozorgasl and Hao Chen.
\newblock Wav-kan: Wavelet kolmogorov-arnold networks, 2024.
\newblock URL \url{https://arxiv.org/abs/2405.12832}.

\bibitem[Zhang et~al.(2025)Zhang, Fan, Cai, and
  Wang]{zhang2025kolmogorovarnoldfouriernetworks}
Jusheng Zhang, Yijia Fan, Kaitong Cai, and Keze Wang.
\newblock Kolmogorov-arnold fourier networks, 2025.
\newblock URL \url{https://arxiv.org/abs/2502.06018}.

\bibitem[Kolmogorov(1957)]{kolmogorov1957representation}
A.~N. Kolmogorov.
\newblock On the representation of continuous functions of many variables by
  superposition of continuous functions of one variable and addition.
\newblock \emph{Dokl. Akad. Nauk SSSR}, 114\penalty0 (5):\penalty0 953--956,
  1957.
\newblock URL \url{http://mi.mathnet.ru/dan22050}.
\newblock MR0111809, Zbl 0090.27103.

\bibitem[Givental et~al.(2009)Givental, Khesin, Marsden, Varchenko, Vassiliev,
  Viro, and Zakalyukin]{Arnold2009}
Alexander~B. Givental, Boris~A. Khesin, Jerrold~E. Marsden, Alexander~N.
  Varchenko, Victor~A. Vassiliev, Oleg~Ya. Viro, and Vladimir~M. Zakalyukin,
  editors.
\newblock \emph{On functions of three variables}, pages 5--8.
\newblock Springer Berlin Heidelberg, Berlin, Heidelberg, 2009.
\newblock ISBN 978-3-642-01742-1.
\newblock \doi{10.1007/978-3-642-01742-1_2}.
\newblock URL \url{https://doi.org/10.1007/978-3-642-01742-1_2}.

\bibitem[de~Boor(2001)]{deboor2001practical}
Carl de~Boor.
\newblock \emph{A Practical Guide to Splines}.
\newblock Springer, New York, revised edition, 2001.
\newblock ISBN 978-0387953663.

\bibitem[Schumaker(2007)]{Schumaker_2007}
Larry Schumaker.
\newblock \emph{Spline Functions: Basic Theory}.
\newblock Cambridge Mathematical Library. Cambridge University Press, 3
  edition, 2007.

\bibitem[Wen et~al.(2016)Wen, Wu, Wang, Chen, and
  Li]{wen2016learningstructuredsparsitydeep}
Wei Wen, Chunpeng Wu, Yandan Wang, Yiran Chen, and Hai Li.
\newblock Learning structured sparsity in deep neural networks, 2016.
\newblock URL \url{https://arxiv.org/abs/1608.03665}.

\bibitem[Yousufi et~al.(2019)Yousufi, Amir, Javed, Tayyib, Abdullah, Ullah,
  Qureshi, Alimgeer, Akram, and Khan]{aplhagasjia}
Musyyab Yousufi, Muhammad Amir, Umer Javed, Muhammad Tayyib, Suheel Abdullah,
  Hayat Ullah, Ijaz~Mansoor Qureshi, Khurram~Saleem Alimgeer, Muhammad~Waseem
  Akram, and Khan~Bahadar Khan.
\newblock Application of compressive sensing to ultrasound images: A review.
\newblock \emph{BioMed Research International}, 2019\penalty0 (1):\penalty0
  7861651, 2019.
\newblock \doi{https://doi.org/10.1155/2019/7861651}.
\newblock URL
  \url{https://onlinelibrary.wiley.com/doi/abs/10.1155/2019/7861651}.

\bibitem[Tian et~al.(2014)Tian, Mansour, Knyazev, and Vetro]{Tian_2014}
Dong Tian, Hassan Mansour, Andrew Knyazev, and Anthony Vetro.
\newblock Chebyshev and conjugate gradient filters for graph image denoising.
\newblock In \emph{2014 IEEE International Conference on Multimedia and Expo
  Workshops (ICMEW)}, page 1–6. IEEE, 2014.
\newblock \doi{10.1109/icmew.2014.6890711}.
\newblock URL \url{http://dx.doi.org/10.1109/ICMEW.2014.6890711}.

\bibitem[SS et~al.(2024)SS, AR, R, and
  KP]{ss2024chebyshevpolynomialbasedkolmogorovarnoldnetworks}
Sidharth SS, Keerthana AR, Gokul R, and Anas KP.
\newblock Chebyshev polynomial-based kolmogorov-arnold networks: An efficient
  architecture for nonlinear function approximation, 2024.
\newblock URL \url{https://arxiv.org/abs/2405.07200}.

\bibitem[Li(2024)]{li2024fastkan}
Ziyao Li.
\newblock Kolmogorov-arnold networks are radial basis function networks.
\newblock \url{https://arxiv.org/abs/2405.06721}, 2024.
\newblock URL \url{https://github.com/ZiyaoLi/fast-kan}.
\newblock arXiv:2405.06721 [cs.LG].

\bibitem[Ta(2024)]{ta2024bsrbfkancombinationbsplinesradial}
Hoang-Thang Ta.
\newblock Bsrbf-kan: A combination of b-splines and radial basis functions in
  kolmogorov-arnold networks, 2024.
\newblock URL \url{https://arxiv.org/abs/2406.11173}.

\bibitem[Howard et~al.(2024)Howard, Jacob, Murphy, Heinlein, and
  Stinis]{howard2024finitebasiskolmogorovarnoldnetworks}
Amanda~A. Howard, Bruno Jacob, Sarah~H. Murphy, Alexander Heinlein, and Panos
  Stinis.
\newblock Finite basis kolmogorov-arnold networks: domain decomposition for
  data-driven and physics-informed problems, 2024.
\newblock URL \url{https://arxiv.org/abs/2406.19662}.

\bibitem[Jost(2013)]{jost2013riemannian}
J.~Jost.
\newblock \emph{Riemannian Geometry and Geometric Analysis}.
\newblock Universitext. Springer Berlin Heidelberg, 2013.
\newblock ISBN 9783662223857.
\newblock URL \url{https://books.google.co.uk/books?id=VRz2CAAAQBAJ}.

\bibitem[Aghaei(2024)]{aghaei2024fkanfractionalkolmogorovarnoldnetworks}
Alireza~Afzal Aghaei.
\newblock fkan: Fractional kolmogorov-arnold networks with trainable jacobi
  basis functions, 2024.
\newblock URL \url{https://arxiv.org/abs/2406.07456}.

\bibitem[Ta et~al.(2025)Ta, Thai, Rahman, Sidorov, and
  Gelbukh]{ta2025fckanfunctioncombinationskolmogorovarnold}
Hoang-Thang Ta, Duy-Quy Thai, Abu Bakar~Siddiqur Rahman, Grigori Sidorov, and
  Alexander Gelbukh.
\newblock Fc-kan: Function combinations in kolmogorov-arnold networks, 2025.
\newblock URL \url{https://arxiv.org/abs/2409.01763}.

\bibitem[Oymak et~al.(2019)Oymak, Fabian, Li, and
  Soltanolkotabi]{oymak2019generalization}
Samet Oymak, Zalan Fabian, Mingchen Li, and Mahdi Soltanolkotabi.
\newblock Generalization guarantees for neural networks via harnessing the
  low-rank structure of the jacobian.
\newblock \emph{CoRR}, abs/1906.05392, 2019.
\newblock \doi{10.48550/arXiv.1906.05392}.
\newblock URL \url{https://arxiv.org/abs/1906.05392}.

\bibitem[Idnani et~al.(2023)Idnani, Madan, Goyal, Schwab, and
  Vedantam]{idnani2023dont}
Daksh Idnani, Vivek Madan, Naman Goyal, David~J. Schwab, and Ramakrishna
  Vedantam.
\newblock Don’t forget the nullspace! nullspace occupancy as a mechanism for
  out‑of‑distribution failure.
\newblock In \emph{Proceedings of the International Conference on Learning
  Representations (ICLR)}, 2023.
\newblock URL \url{https://openreview.net/forum?id=39z0zPZ0AvB}.

\bibitem[Coifman and Wickerhauser(1992)]{119732}
R.R. Coifman and M.V. Wickerhauser.
\newblock Entropy-based algorithms for best basis selection.
\newblock \emph{IEEE Transactions on Information Theory}, 38\penalty0
  (2):\penalty0 713--718, 1992.
\newblock \doi{10.1109/18.119732}.

\bibitem[Zhuang and Baras(1994)]{zhuang1994optimal}
Yan Zhuang and John~S. Baras.
\newblock Optimal wavelet basis selection for signal representation.
\newblock Technical research report cshcn t.r. 94‑7 (isr t.r. 94‑3), Center
  for Satellite and Hybrid Communication Networks (CSHCN), Institute for
  Systems Research, University of Maryland, College Park, MD, USA, 1994.
\newblock Available via UMD DRUM API:
  \url{https://api.drum.lib.umd.edu/server/api/core/bitstreams/50447531-f06c-471f-a2d9-1a7cc76d91cf/content}.

\bibitem[Liu et~al.(2024)Liu, Wang, Vaidya, Ruehle, Halverson,
  Solja{\v{c}}i{\'c}, Hou, and Tegmark]{pykan2025}
Ziming Liu, Yixuan Wang, Sachin Vaidya, Fabian Ruehle, James Halverson, Marin
  Solja{\v{c}}i{\'c}, Thomas~Y Hou, and Max Tegmark.
\newblock {pykan: Kolmogorov‑Arnold Networks (KAN)}.
\newblock \url{https://github.com/KindXiaoming/pykan}, 2024.
\newblock GitHub repository, version v0.2.8 (latest release as of
  Nov 14, 2024). MIT License.

\bibitem[Hu(2018)]{hu2018torchdct}
Ziyang Hu.
\newblock torch-dct: Discrete cosine transform (dct) for pytorch.
\newblock GitHub repository, 2018.
\newblock \\url{https://github.com/zh217/torch-dct} (accessed July 30, 2025).

\bibitem[McArthur et~al.(2025)McArthur, Mineo, Poole, Bomphray, and
  Mehnen]{afpcite}
Stig McArthur, Carmelo Mineo, Alastair Poole, Iain Bomphray, and Jörn Mehnen.
\newblock Future-proof adaptive path correction in automated fibre placement: A
  concept demonstration.
\newblock In \emph{2025 IEEE International Conference on Simulation, Modeling,
  and Programming for Autonomous Robots (SIMPAR)}, pages 1--6, 2025.
\newblock \doi{10.1109/SIMPAR62925.2025.10978996}.

\bibitem[Koptelov et~al.(2025)Koptelov, Said, and Tretiak]{KOPTELOV2025112655}
Anatoly Koptelov, Bassam~El Said, and Iryna Tretiak.
\newblock Enhancing afp manufacturing with ai: Defects forecasting and
  classification.
\newblock \emph{Composites Part B: Engineering}, 304:\penalty0 112655, 2025.
\newblock ISSN 1359-8368.
\newblock \doi{https://doi.org/10.1016/j.compositesb.2025.112655}.
\newblock URL
  \url{https://www.sciencedirect.com/science/article/pii/S1359836825005566}.

\end{thebibliography}


\end{document}